\documentclass[12pt,reqno]{amsart}
\usepackage{amssymb}
\usepackage{cases}
\usepackage{bbm}
\usepackage{mathrsfs}
\usepackage{graphicx}
\usepackage{amsfonts}
\usepackage{enumerate}
\usepackage{ulem}
\usepackage{amssymb, amsmath, amsfonts, latexsym}
\usepackage{enumerate}
\usepackage{color}
\usepackage{fmtcount}
\newtheorem{thm}{Theorem}[section]
\newtheorem{cor}[thm]{Corollary}
\newtheorem{lem}[thm]{Lemma}
\newtheorem{prop}[thm]{Proposition}
\newtheorem{example}[thm]{Example}
\newtheorem{remarks}[thm]{Remark}
\newtheorem{defn}[thm]{Definition}
\newtheorem{hyp}[thm]{Hypothesis}
\numberwithin{equation}{section}

\date{}

\newcommand{\ee}{\mathbb{E}}

\newcommand{\nn}{\mathbb{N}}
\newcommand{\rr}{\mathbb{R}}
\newcommand{\pp}{\mathbb{P}}

\newcommand{\argmin}{\operatorname{argmin}}

\def\BB{\mathcal B}

\def\FF{\mathcal F}

\def\HH{\mathcal H}

\def\M{\mathbf M}

\def\vep{\varepsilon}
\def\<{\langle}
\def\>{\rangle}

\def\d"{^{\prime\prime}}
\def\bhyp{\begin{hyp}}
\def\nhyp{\end{hyp}}
\def\bbeq{\begin{equation}}
\def\nneq{\end{equation}}
\def\bdef{\begin{defn}}
\def\ndef{\end{defn}}
\def\bthm{\begin{thm}}
\def\nthm{\end{thm}}
\def\bprop{\begin{prop}}
\def\nprop{\end{prop}}
\def\brmk{\begin{remarks}}
\def\nrmk{\end{remarks}}
\def\bexa{\begin{example}}
\def\nexa{\end{example}}
\def\blem{\begin{lem}}
\def\nlem{\end{lem}}
\def\bcor{\begin{cor}}
\def\ncor{\end{cor}}
\def\bexe{\begin{exe}}
\def\nexe{\end{exe}}
\def\bprf{\begin{proof}}
\def\nprf{\end{proof}}
\def\dsp{\displaystyle}
\def\bdes{\begin{description}}
\def\ndes{\end{description}}

\def\var{{\rm Var}}

\def\vc{{\rm vc}}

\title[Non-asymptotic estimates of the minimal risk]{Non-asymptotic estimates of the minimal risk in statistical learning}
\author{Liming Wu}
\address{Liming Wu.  Laboratoire de Math\'ematiques Blaise Pascal, CNRS-UMR
6620, Universit\'e Clermont-Auvergne (UCA), 63000 Clermont-Ferrand, France.}
\email{li-ming.wu@uca.fr}

\author{Sen Yang}
\address{Sen Yang. IASM, Harbin Institute of Technology, China}
\email{syang8734@gmail.com}

\date{\today}

\begin{document}
\begin{abstract} In this paper we prove some concentration inequalities for two types of error probabilities in the Empirical Risk Principle (ERP) in statistical learning, which provide a lower bound  and an upper bound of the minimal risk (in terms of the minimal empirical risk) with non-asymptotic high confidence. The usual boundedness condition of the empirical risk function is relaxed to the Gaussian or exponential integrability condition. The confidence of the lower bound of the minimal risk is shown to be independent of the number of training parameters and the dimension of the input vectors, signifying that we can  quickly verify the deficiency of a learning machine; and the confidence of the upper bound of the minimal risk is proved to be high provided that the sample size $n$ is much greater than the box dimension of the parameter set $\Theta$ in the Orlicz metric
$d_{\psi_1}$ associated with the risk functions. 
Our work is based on  Talagrand's concentration inequalities (the sharp versions by Bousquet and Klein-Rio), transport-entropy inequalities and the recent progress in the theory of empirical processes and statistical learning. 
\end{abstract}
\maketitle



\section{Introduction}
\subsection{Two error probabilities in  Empirical Risk Principle (ERP) in statistical learning.}

Learning machines (such as (deep) neural networks) are furnished with a special class of functions
$$
\FF =\{f(z, \theta); \theta\in \Theta\}
$$
where $\Theta\subset \rr^N$ is a domain of $\rr^N$, $N$ being the number of training parameters which is often huge\footnote{$N\asymp 10^{9}$ to $10^{13}$}, and 
\begin{enumerate}[(a)]
  \item in the regression problem where $y=f(x)+\eta$ where $\eta$ is the centered noise, $z=(x,y)$ and $f(z,\theta)=f(x,\theta)$ is designed to approximate the unknown function $f(x)$; 
  \item in the maximum likelihood problem $f(z,\theta)$ is designed to approximate the likelihood $g(z)$ or the log-likelihood $\log g(z)$  of the (unknown) distribution $\mu(dz)=g(z) dz$ of the state variable (high dimensional vector) $Z$.
\end{enumerate}

To quantify what is meant by the ``best way'', introduce a risk (or loss) function
$$
Q(z,\theta)=\begin{cases} |y-f(x, \theta)|^2  \ &\text{ or } \\
|y-f(x,\theta)| \ &\text{ both in the regression problem} \\
-\log f(z,\theta) &\text{ in the Maximum Likelihood problem} \\
&\text{ or others}
\end{cases}
$$
The true (but unknown) risk for a given $\theta$ is
\bbeq\label{Eq11}
R(\theta)=\ee Q(Z,\theta),
\nneq
 and its infimum $R_*:=\inf_{\theta} R(\theta)$ characterizes the theoretical minimal risk of the learning machine.

When $Q(z,\theta)=|y-f(x, \theta)|^2$ and $Z=(X,Y)$ in the (nonlinear) regression, the theoretical risk for a given $\theta$ is
$$
R(\theta)= \ee |Y- f(X,\theta)|^2 = \ee |Y-f_0(X)|^2 + \ee |f_0(X)-f(X,\theta)|^2
$$
where $f_0(x)=\ee(Y|X=x)$ is the conditional expectation, known as the non-linear regression function. Then the theoretical minimal risk of the learning machine is
\begin{equation}
\inf_{\theta\in\Theta} R(\theta) = \ee |Y-f_0(X)|^2 + \inf_{\theta\in\Theta} \ee |f_0(X)-f(X,\theta)|^2.
\end{equation}
The first term at the right hand side (r.h.s.) cannot be diminished by any learning machine (because of the "random" dependence assumption of $Y$ upon $X$), and the least-square error
$$\inf_{\theta\in\Theta} \ee |f_0(X)-f(X,\theta)|^2.$$
quantifies the theoretical efficiency of the learning machine. It is known today that when $X$ is of uniform distribution on $[0,1]^d$ and $f_0$ is smooth enough,  if the neural network is sufficiently wide and/or deep, $\dsp \inf_{\theta\in\Theta} \ee |f_0(X)-f(X,\theta)|^2$ can be made arbitrarily small (known as universal approximation theorem); see \cite{E23ICM, SSS25} and the references therein.

One main purpose of machine learning  is to minimize the empirical risk function
\begin{equation}
R_{E, n}(\theta ) = \frac{1}{n} \sum_{k=1}^{n} Q(Z_k, \theta)
\end{equation}
among all $\theta \in \Theta$, i.e. to find the minimizers of
\begin{equation}
\arg\min_{\theta \in \Theta} R_{E, n}(\theta )=\{\hat{\theta}_n \in \Theta\mid R_{E, n}(\hat{\theta}_n)\leqslant  R_{E, n}(\theta ), \ \forall \theta \in \Theta\}.
\end{equation}
This is called ''training the parameters" in machine learning.  Here $R_{E,n}$ is the empirical risk, whereas $R$ is the unknown true risk. In the language of statistical learning theory, the basic problem is to quantify  {\it  how close the minimal empirical risk $\inf_{\theta}R_{E, n}(\theta )$ is  to  the true minimal risk
$\inf_{\theta} R(\theta)$.}

 By the law of large numbers,
$$
R_{E,n}(\theta)\to R(\theta):=\ee Q(Z, \theta)
$$
where the true risk $R(\theta)$ is usually unknown.
 
\medskip
{\bf Empirical Risk Principle} ({\bf ERP} in short), laid by Vapnik \cite{Vapnik} as a basic first principle for statistical learning theory, means that
\begin{equation}\label{ERP}
\aligned
  p_{+}(n,\vep)&:=\pp\left( \inf_{\theta\in\Theta} R(\theta)> \inf_{\theta\in\Theta} R_{E,n}(\theta)+ \vep \right)\\
   p_{-}(n,\vep)&:=\pp\left(\inf_{\theta\in\Theta} R(\theta)< \inf_{\theta\in\Theta} R_{E,n}(\theta)-\vep \right)\\
\endaligned
\end{equation}
both tend to zero for any $\vep>0$. In the classical approach, this is deduced from (uniform) law of large numbers for empirical processes, namely from the control of
$$
\sup_{\theta\in\Theta}|R_{E,n}(\theta)-R(\theta)|.
$$
This is exactly the point of view of the Glivenko--Cantelli theory developed in the classic theory of empirical processes \cite{Vapnik, vdVW96} and it is a theoretical basis of statistical learning \cite{Kol11, BLM13, Wainwright19}. When $|Q(z,\theta)|\leqslant M$ is bounded, a necessary and sufficient condition for the Glivenko--Cantelli theorem in terms of the VC entropy number is known (\cite[\S 2.3.4, Theorem 2.3]{Vapnik}).

The first error probability $p_+(n,\vep)$ gives an upper bound of the theoretical minimal risk:
$$\pp\left(\inf_{\theta\in\Theta} R(\theta) \leqslant  \inf_{\theta\in\Theta} R_{E,n}(\theta) + \vep\right) = 1-p_+(n,\vep)$$
($1-p_+(n,\vep)$ is the so-called confidence level);  whereas the second error probability $p_-(n, \vep)$ gives a lower bound of the theoretical minimal risk:
$$\pp\left(\inf_{\theta\in\Theta} R(\theta) \geqslant  \inf_{\theta\in\Theta} R_{E,n}(\theta) - \vep\right) = 1-p_-(n, \vep).$$
In other words, $p_+(n,\vep)$ quantifies the efficiency of a learning machine;  $p_-(n,\vep)$ quantifies the non-efficiency or the deficiency of a learning machine, both based on the minimal empirical risk $\inf_{\theta\in\Theta} R_{E,n}(\theta)$. Thus, the control of $p_+(n,\vep)$ and $p_-(n,\vep)$ may be viewed as a quantitative form of the generalization problem at the level of minimal risks.
Estimating the two error probabilities $p_{+}(n, \vep)$ and $p_{-}(n, \vep)$ is then a basic question in statistical learning. 

In modern learning problems, models often have very large parameter sets while the available dataset remains finite; this combination can lead to overfitting \cite[p. 7]{SSS25}. Therefore quantitative non-asymptotic estimates become essential. However, the classical limit theorems such as Donsker's invariance principle (i.e. uniform central limit theorem, see  \cite{LT91}, \cite{vdVW96}), the large and moderate deviation principles (\cite{Wu94emp}, \cite{WWW10}), which are only asymptotic (when $n\to +\infty$), cannot be applied directly, because the available sample size $n$ cannot be much greater than the number $N$ of parameters in practice, and the dimension $d$ of the input vector $Z$ is often very high (for example, a photo with $256\times256$ pixels).

For non-asymptotic estimates of the error probabilities, there has been substantial progress both in probability and in statistics in the last thirty years, largely initiated by Talagrand's concentration inequalities
\cite{Talagrand94AP, Talagrand_IHES95, Tal96b}.

\subsection{Bousquet's and Klein-Rio's sharp versions of Talagrand's concentration inequality}

Talagrand (\cite{Talagrand94AP, Talagrand_IHES95, Tal96b}) investigated in depth the concentration phenomena on product measure spaces and renewed the theory of empirical processes.
Ledoux \cite{Ledoux96} (see also his lecture course \cite{Ledoux99} for a pedagogic and systematic study) gave a much simpler  proof of Talagrand's concentration inequality by means of the log-Sobolev inequality, and the proof of Ledoux was  refined considerably by  Massart \cite{Massart00AP} for finding explicit constants in Talagrand's concentration inequality. The following versions with sharp constants  of Talagrand's concentration inequality are due to Bousquet \cite[Theorem 2.11]{Bousquet02} and Klein-Rio \cite{KR_AOP05} 
(see also the books \cite{Kol11, BLM13}).

\bthm\label{Tala_concentration} {\bf (Bousquet's inequality)} Given
\begin{enumerate}
  \item a sequence of i.i.d. r.v.'s $(Z_k)_{k\geqslant 1}$ valued in some Polish space $S$ equipped with the Borel $\sigma$-field, of common law $\mu$;
  \item an at most countable class $\HH$ of upper bounded measurable functions $h$ on $S$ such that $h-\mu(h)\leqslant b$ (resp. $|h-\mu(h)|\leqslant b$) for some positive constant $b$;
\end{enumerate}
let
\bbeq\label{EmpM}
L_n=\frac{1}{n} \sum_{k=1}^{n} \delta_{Z_k}
\nneq
be the empirical distribution of the i.i.d. sample $Z_i, 1\leqslant i\leqslant n$ ($\delta_z$ being the Dirac measure at point $z$),
$$
U = \sup_{h\in\HH} \frac{1}{n}\sum_{k=1}^{n} (h(Z_k)-\ee h(Z))= \sup_{h\in\HH} (L_n(h)-\mu(h))\ \text{ (resp. }\ U = \sup_{h\in\HH} \left|\frac{1}{n}\sum_{k=1}^{n} [h(Z_k)- \mu(h)]\right|\;)
$$
and
\bbeq\label{3thm_TM1}
\sigma^2(\mathcal{H})=\sup_{h\in\mathcal{H}} \var_{\mu}(h).
\nneq
Then for any $n\geqslant 1$, $x>0$, 
\bbeq\label{3thm_TM2}
\pp\left(U > \ee U  + \sqrt{\frac{ 2 \left( \sigma^2(\mathcal{H}) +  2 b \ee U \right) x}{n}} +  \frac{bx}{3n}\right)\leqslant e^{-x}.
\nneq
\nthm

\bthm\label{thm_KR} {\bf (Klein-Rio's inequality)} In the framework of Theorem \ref{Tala_concentration}, assume that $|h|\leqslant b$, for the two types suprema $U$  of empirical processes given in Theorem \ref{Tala_concentration},
\bbeq\label{thm_KR1}
\pp\left(U < \ee U  -\left[ \sqrt{\frac{ 2 \left( \sigma^2(\mathcal{H}) +  2 b \ee U \right) x}{n}} +  \frac{bx}{n}\right]\right)\leqslant e^{-x}, \qquad \forall x>0.
\nneq

\nthm

Applying it to $\HH=\{Q(z, \theta)-R(\theta); \theta\in \Theta\}$ and assuming that $Q(z,\theta)- \ee Q(Z,\theta)$ is bounded in absolute value by $L$, separable in $\theta$ (for circumventing the countable condition in the theorem above), we get (in a simple and rough way)
$$
\aligned
\max\{p_+(n, \vep), p_-(n,\vep)\}&\leqslant \pp\left(|\inf_{\theta\in\Theta} R_{E,n}(\theta)-\inf_{\theta\in\Theta} R(\theta)| > \vep \right)\\
& \leqslant \pp\left(\sup_{\theta\in\Theta} |R_{E,n}(\theta)-R(\theta)| > \vep \right)\\
& \leqslant e^{-x}, \qquad x>0
\endaligned
$$
where
\bbeq
\vep=\vep(n, x)=\ee U  + \sqrt{\frac{ 2 \left( \sigma^2(\mathcal{H}) +  2 L \ee U \right) x}{n}} +  \frac{L x}{3n}
\nneq
and $U=\sup_{\theta\in \Theta} |R_{E,n}(\theta)-R(\theta)|$. Except the bias 
$$\ee \sup_{\theta\in \Theta} |R_{E,n}(\theta)-R(\theta)|, $$ the other terms controlling the random fluctuation of $\sup_{\theta\in\Theta} |R_{E,n}(\theta)-R(\theta)|$,
are independent of $d$ and $N$ (depending only upon the maximal variance $\sigma^2(\Theta)$).

For further generalizations of Talagrand's inequality to the non-i.i.d. case, see Klein and Rio \cite{KR_AOP05}; to the unbounded case, see Adamczak \cite{Ada08}. See the excellent book \cite{BLM13} by Boucheron, Lugosi and Massart for the developments and applications of Talagrand's concentration  inequalities in many areas.

\subsection{Estimate of the bias in terms of the VC dimension}

Let ${\rm vc}(\HH)$ be the VC (Vapnik-Chervonenkis) dimension $m={\rm vc}(\HH)$ of $\HH=\{Q(z, \theta); \theta\in \Theta\}$ (see \cite{vdVW96, Vapnik, Vershynin20, Wainwright19}). The following concentration inequality in Vapnik \cite[\S3.1, Theorem 3.1]{Vapnik} is well known to specialists:
$$
\ee \sup_{\theta\in \Theta} |R_{E,n}(\theta)-R(\theta)|\leqslant C \sqrt{\frac{m (\log \frac{n}{m}+1)}{n}}
$$
for some explicit absolute constant $C$, when $|Q(z,\theta)|\leqslant 1$.

This is considerably improved in

\bthm[{\cite[Theorem 13.7]{BLM13}, \cite[Theorem 8.3.23]{Vershynin20}}]\label{3thm_VCbias} Assume that $a\leqslant h\leqslant b$ for all $h\in\mathcal{H}$ and the VC dimension ${\rm vc}(\mathcal{H})$ of $\HH$ is finite (the so-called VC class). Then
\begin{equation}\label{3thm_VCbias1}
  \ee \sup_{h \in \mathcal{H}}|L_n(h)-\mu(h)| \leqslant K \sqrt{\frac{{\rm vc}(\mathcal{H})}{n}}(b-a).
\end{equation}
where $K>0$ is an absolute constant (see \cite[Theorem 13.7]{BLM13} for explicit constants version).
\nthm

It implies, together with Bousquet's concentration inequality, that when $|Q(Z,\theta)|, \theta\in \Theta$ are bounded by some constant,  if the sample size
\bbeq\label{1Eq11}
n \succeq \frac{1}{\vep^2} {\rm vc}(\HH),
\nneq
the minimal empirical risk $\inf_{\theta} R_{E,n}(\theta)$ will approach the theoretical minimal risk $\inf_{\theta} R(\theta)$ within precision $\vep$ with quantitative high probability.

In other words, this bias estimate transforms the dependence on $N,d$ into the dependence upon the VC dimension of the risk functions  $\HH=\{Q(z, \theta); \theta\in \Theta\}$.

For the neural networks, tight estimates of the maximal VC dimension of $\{f(z,\theta); \theta\in \Theta\}$ in neural networks were obtained recently by  Bartlett {\it et al.} in \cite{BH19_vcdim} after a series of efforts (see \cite{BH19_vcdim} for related references): the VC dimension is $\Theta(NU)$, where $U$ is the number of nonlinear units, and is between $\Omega(NL\log(N/L))$ and $O(NL\log N)$, where $L$ is the number of layers.  Although VC-dimension bounds are classical, they are not necessarily informative in the deep-learning setting, and closing the gap between theory and empirical evidence remains an active area of research \cite[p. 390]{SSS25}.

\subsection{Some recent progress}

 Recent developments in statistical learning theory indicate that a central issue is to obtain quantitative non-asymptotic bounds and to understand the localized phenomena associated with empirical minimization.

Besides the classical VC dimension, an important development in modern statistical learning theory is to complement distribution-free complexity measures by data-dependent quantities. Bartlett and Mendelson \cite{BM02} gave a systematic treatment of Rademacher and Gaussian complexities as data-dependent measures of function-class complexity, establishing risk bounds and structural results for model classes, while Bartlett, Bousquet and Mendelson \cite{BBM05} showed that local Rademacher complexities, based on a subset of functions with small empirical error, may yield much sharper bounds. Those results explain why, for ERP, one should not only study the whole class $\HH=\{Q(z,\theta);\theta\in\Theta\}$ globally, but also the localized part selected by empirical minimization.

Another line of work is devoted to a direct study of the empirical risk minimization (ERM) algorithm. Bartlett and Mendelson \cite{BM06} showed that a direct analysis of the ERM algorithm yields significantly better bounds than those obtained from global comparisons, and that these bounds are essentially sharp, while Mendelson \cite{Men08} established lower bounds showing that, without further assumptions, the uniform error rate of ERM cannot in general be faster than $1/\sqrt{n}$, where $n$ is the sample size. Bousquet--Elisseeff \cite{BE02} developed the stability approach to obtaining generalization bounds. Recently, Feldman and Vondrak \cite{FeldmanV18, FeldmanV19},  Bousquet {\it et al.} \cite{BousquetKZ20} obtained sharp bounds of uniformly stable algorithms, improved later by Klochkov and Zhivotovskiy \cite{KlochkovZ21} when  the  theoretical risk $R$ is strictly convex in $\theta$.  
Escande \cite{Esc23} studied concentration inequalities for the distance between the set of empirical minimizers and the set of theoretical minimizers, rather than only the excess risk. Related localization and oracle-inequality phenomena for empirical risk minimization were studied by Koltchinskii \cite{Kol06}, who derived excess-risk bounds in terms of localized empirical and Rademacher complexities. These works confirm that, beyond asymptotic limit theorems, one needs quantitative non-asymptotic estimates for the deviations of the minimal empirical risk, and, if possible, for the minimizers themselves.

For a systematic and pedagogic account of oracle inequalities and excess-risk bounds in empirical risk minimization, we refer to the lecture notes of Koltchinskii \cite{Kol11} at the {\it Ecole d'Et\'e de Probabilit\'es de St Flour}.

 Recent developments in high-dimensional probability show that the error probabilities depend often on the dimension $d$ and the number  $N$ of parameters. For example if one uses the $L^1$-Wasserstein distance $W_1$, Fournier and Guillin \cite{FG15} (2015)  showed that if the sample $Z_k$ is valued in $\rr^d$,
\bbeq\label{1SW1}
W_1(L_n, \mu) \asymp \frac{1}{n^{1/d}}, \qquad \text{ if } d\geqslant 3
\nneq
with high probability under fairly general conditions, where
$$
L_n=\frac{1}{n} \sum_{k=1}^{n} \delta_{Z_k} \ \text{ ($\delta_\cdot$ being the Dirac point measure at point $\cdot$)}
$$
is the empirical distribution of the sample. Their result generalizes a discovery of M. Ajtai, J. Koml\'os, G. Tusn\'ady  \cite{AKT84} for  the uniform distribution $\mu$ on $[0,1]^{d}$. This result has recently attracted much attention,  especially in the critical case $d=2$. See Ambrosio {\it et al.} \cite{AST19},  Talagrand \cite{Talagrand18} in the i.i.d. case, Wang \cite{WangFY23jems} in the diffusion case and the references therein. See the book \cite{Vershynin20} (2018) by Vershynin  on high-dimensional probability for a state-of-the-art account and references therein.

 For the theory and applications of statistical learning to deep learning and neural networks, the reader is referred to the books \cite{Wainwright19} (2019) by Wainwright, \cite{Bach24} (2024) by Bach, \cite{SSS25} (2025) by Spiliopoulos, Sowers and Sirignano and references therein.

\subsection{Purposes of this work.}

The lower-bound problem $p_{-}(n,\varepsilon)$ and the upper-bound problem $p_{+}(n,\varepsilon)$ are of different natures. While the former can often be reduced to a concentration estimate for a single (suitably chosen) observable, the latter requires a more delicate localized control of the empirical-risk fluctuations along suitable risk layers.
The purpose of this paper is threefold:

\begin{enumerate}
  \item to establish a dimension-free sharp Bernstein's concentration inequality for $p_{-}(n, \vep)$; and as an application we show that we can verify the non-efficiency of a learning machine with a sample size $n$ that is independent of $N$ and $d$ and need not be very large (to the best of our knowledge, this is not discussed in the current literature); 
  \item to remove the boundedness assumption, which is indispensable, for instance, when
      the noise is Gaussian;
  \item to present some estimates of $p_{+}(n,\vep)$ ensuring the efficiency of a learning machine, in particular by replacing the VC dimension $\vc(\HH)$ by a distribution-dependent metric entropy, and the associated box-dimension, generalizing slightly the known results presented in \cite{Kol11, BLM13, Wainwright19} to the unbounded case.
  Here we use the Bernstein concentration inequalities as in \cite{BLM13}, instead of the symmetrization technique and the classical results on Rademacher sequences in the theory of empirical processes (\cite{vdVW96}) or in statistical learning theory \cite{BM02, BBM05, Kol11, BLM13, Wainwright19}. The proof of the upper bound is based on a risk-level localization (peeling argument) of the empirical process.

\end{enumerate}

\subsection{Organization.}

The paper is organized as follows. In Section 2, we develop non-asymptotic dimension-free estimates for $p_-(n,\vep)$, which yield lower bounds for the theoretical minimal risk via Bernstein concentration inequalities for the empirical risk: the main new contribution here is a Bernstein's concentration inequality with sharp constant without boundedness. Section 3 is devoted to upper bounds in terms of the $d_{\psi_1}$-metric entropy and risk-level localization, yielding non-asymptotic estimates of $p_+(n,\vep)$ under suitable entropy assumptions. In Sections 4 and 5, we present the proofs of several main theorems. Several technical lemmas are collected in the Appendix.

\bigskip
{\bf Notations. }
\begin{itemize}
  \item For $x,y\in \rr^d$, $|x|$ is the Euclidean norm, $x\cdot y=\<x,y\>$ is the Euclidean inner product.

  \item $I_d$ is the identity matrix of size $d\times d$.

  \item $\mu(f)=\int fd\mu$ for a function $f$ and a measure $\mu$ on the same measurable space.
  \item for two sequences of positive numbers $a_n, b_n$,
  \begin{enumerate}
    \item $a_n \succeq b_n$ or $b_n \preceq a_n$ means that $a_n\geqslant C b_n$ for some {\it big} constant $C$ for all $n\geqslant n_0$;
    \item $a_n \gtrsim b_n$ or $b_n \lesssim a_n$ means that $a_n\geqslant C b_n$ for some constant $C$ for all $n$;
    \item $a_n \asymp b_n$ means that $C^{-1} b_n\leqslant a_n\leqslant C b_n$ for some constant $C$.
  \end{enumerate}
  \item the constant $C$ may change from one line to another, only used in Corollaries or Remarks for explaining the main results.  
\end{itemize}

\section{ Bernstein-type concentration inequalities and lower bounds of the risk}

\subsection{Framework}
We will work in  the mathematical framework of the theory of empirical processes, but motivated by the problems of statistical learning. The following assumption will be imposed throughout the paper.

\medskip
{\bf (H1)} {\it

\begin{enumerate}
  \item[(a)] $(Z_1,\cdots, Z_n)$ is a sample of i.i.d. random variables valued in some Polish space $S$ equipped with the Borel $\sigma$-field $\BB$, of distribution $\mu$ of some $S$-valued random variable $Z$, defined on some complete probability space $(\Omega,\FF,\pp)$; and the empirical distribution of the sample is
  $$
  L_n=\frac{1}{n} \sum_{k=1}^{n} \delta_{Z_k};
  $$
  ($L_n$ is a random element of the space $M_1(S)$ of probability measures on $(S, \BB)$.)
  \item[(b)] a non-empty subset $\Theta\subset \rr^N$;
  
  \item[(c)] a family of measurable functions $\HH=
\{Q_\theta:S\to\rr:
Q_\theta(z)=Q(z,\theta),\ \theta\in\Theta\}$ (interpreted as risk functions), contained in $L^2(S, \BB, \mu)=:L^2(\mu)$, such that
  $(Q_\theta:=Q(z,\theta))_{\theta \in \Theta}$, regarded as stochastic process indexed by $\theta\in \Theta$ defined on the probability space $(S,\BB,\mu)$,
  is separable in the sense of Doob (see \cite{vdVW96} for definition).

\end{enumerate}
}

Here we remove the boundedness assumption on the risk functions $Q_\theta$. In reality when learning the log-likelihood function $\log g(z)$, the risk functions are usually unbounded, as well as in the linear or non-linear regression problems when $X,Y$ are unbounded.

Under the separability assumption in {\bf (H1)}, the empirical risk function
  $$
  R_{E,n}(\theta)= \frac{1}{n} \sum_{k=1}^{n} Q(Z_k, \theta), \qquad \theta\in\Theta
  $$
 regarded as stochastic process indexed by $\theta\in \Theta$, is again separable.

We denote by
\bbeq\label{2Eq1}
\sigma^2(\Theta)= \sup_{\theta\in \Theta} \sigma^2(\theta),\qquad \sigma^2(\theta):=\var_\mu(Q(\cdot, \theta)) =  \int_S Q(z,\theta)^2 \mu(dz) - R(\theta)^2,
\nneq
the maximal variance of $\HH$, where $R(\theta)= \ee Q(Z,\theta)=\int_S Q(z,\theta)\mu(dz)$ is the theoretical risk for the given parameter $\theta$.

\subsection{Some known results about Bernstein's concentration inequality}
We begin with the well-known Bernstein inequalities.  Let
\bbeq\label{entropy1}
H(\nu|\mu) = \begin{cases}\int_S h \log h d\mu; &\text{ if } \nu=h\mu\\
+\infty,  &\text{ otherwise}
\end{cases}
\nneq
be the relative entropy of $\nu\in M_1(S)$ w.r.t. $\mu$.

We begin with a particular case of a general result of Gozlan and L\'eonard \cite{GL07} (see also their survey \cite{GL10}), relating the concentration inequalities with the transport-entropy inequality.

\bthm\label{thm_GL} {\bf (Gozlan-L\'eonard \cite{GL07})} Given the constants $c_B>0, M\geqslant  0$ and a $\mu$ exponentially integrable function $f$ on $S$, i.e.
\bbeq\label{ExpInteg}
\exists \delta>0 :\ \int_S e^{\delta |f|} d\mu <+\infty,
\nneq
the following properties are equivalent:
\begin{enumerate}
  \item The log-Laplace transform of $f(X)$ satisfies
  \bbeq \label{Bern-a}
  \Lambda(\lambda):=\log \ee e^{\lambda[f(X)-\mu(f)]} \leqslant \frac{c_B\lambda^2
  }{2(1-\lambda M)}, \qquad \lambda\in (0,1/M);\nneq

  \item for any
  $r>0$ and $n\geqslant  1$,
  \bbeq \label{Bern-b}
  \pp\left(L_n(f)-\mu(f)
  > r\right)\leqslant \exp\left(-n
  \frac{2r^2}{c_B \left(\sqrt{1+\frac{2Mr}{c_B}}+1\right)^2}\right),
  \qquad r>0 ;
  \nneq

  \item for any $x>0$ and $n\geqslant  1$,
  \bbeq \label{Bern-c}
  \pp\left(L_n(f)-\mu(f)>
  \sqrt{\frac{2c_B x}{n}}+ M\frac{x}{n}\right)\leqslant e^{-x};
  \nneq

  \item the following transport-entropy inequality holds:
  \bbeq\label{TE1}
  \nu(f)- \mu(f) \leqslant \sqrt{2 c_B H(\nu|\mu)} + M H(\nu|\mu),
  \nneq
  for all $\nu\in M_1(S)$ such that $\nu\ll \mu$ and $\nu(|f|)<+\infty$.
\end{enumerate}
In particular, when \eqref{Bern-a} holds, then the normalized sum $$W_n(f)=\frac{1}{\sqrt{n}}\sum_{k=1}^{n} (f(Z_k)-\mu(f))$$ satisfies
 \bbeq \label{Bern-a2}
  \log \ee e^{\lambda W_n(f)} \leqslant \frac{c_B\lambda^2
  }{2(1-\lambda M/\sqrt{n})}, \qquad \lambda\in (0, \sqrt{n}/M), 
  \nneq
and the following Bernstein's concentration inequality holds:
\bbeq \label{Bern-d} \pp\left(W_n(f)>
r\right)\leqslant \exp\left(- \frac{r^2}{2(c_B +(M/\sqrt{n})r)}\right), \qquad r>0.
\nneq
\nthm

The usual version of Bernstein's inequality is \eqref{Bern-d}, the refined versions \eqref{Bern-b},  \eqref{Bern-c} are more commonly encountered in research papers.

By the second order limit expansion of Taylor-Young at $\lambda=0+$ in \eqref{Bern-a}, we see that the Bernstein concentration constant $c_B$ satisfies
\bbeq
c_B \geqslant \var_\mu(f)=\mu(f^2)-(\mu(f))^2.
\nneq
If we don't care about the estimates of involved constants $c_B, M$ (just for their existence), it is well known (\cite{Vershynin20}) that each of the inequalities above in Theorem \ref{thm_GL} for both $f$ and $-f$ (i.e. two sided Bernstein's inequality) is equivalent to
the exponential integrability condition \eqref{ExpInteg} of $f$.

Notice also that when $M=0$, the Bernstein concentration inequality becomes Hoeffding's Gaussian concentration inequality for which the existence of $c_B$ for both $\pm f$ is equivalent to the Gaussian integrability of $f(X)$ (\cite{DGW04}).

A first problem addressed in this work is to determine the constants $c_B, M$ in \eqref{Bern-a}, so that $c_B$ is close to  the optimal one: $\var_\mu(f)$. When $c_B$ is arbitrarily close to
$\var_\mu(f)$, the Bernstein concentration inequality becomes sharp in moderate deviations (\cite{dPenaLaiShao09}, \cite{Rio00}):

\bbeq\label{MD1}
\lim_{n\to+\infty} \frac{1}{a(n)^2} \log \pp \left(\frac{\sqrt{n}}{a(n)} \left(L_n(f)-\mu(f)\right)>
r\right)=-\frac{r^2}{2\var_\mu(f)}, \ r>0
\nneq
for any given sequence of positive numbers $(a(n))_{n\in \nn}$ such that $1\ll a(n)\ll \sqrt{n}$.

Such a sharp Bernstein's inequality is well known
if $f$ is bounded   or just upper bounded (\cite{dPenaLaiShao09}, \cite{Rio00}, \cite{Vershynin20}):

\bthm\label{thmA} If $f\leqslant b$ is upper bounded and $\mu$-square integrable  where $b> 0$, \eqref{Bern-a} holds with
  \bbeq \label{3thmBern2a}
  c_B= \mu(f^2),\qquad  M=\frac{b}{3}.
  \nneq
\nthm
It becomes sharp when $\mu(f)=0$ because $c_B=\mu(f^2)=\var_\mu(f)$.

Usually one proves the Bernstein concentration inequality via the control of Laplace transform as in the above argument. Completely different from this classic method of Laplace transform, Bolley-Villani \cite{BolleyVillani05}, generalizing Djellout {\it et al.} \cite{DGW04}, proved the corresponding transport-entropy inequality (weighted CKP inequality), and their constants are improved by Gozlan-L\'eonard \cite{GL07}. For stating their result, recall at first the Orlicz norm of a measurable function $f$ on $S$,  associated with the convex function
\bbeq
\psi_p(x) = e^{|x|^p}-1
\nneq
where $p\geqslant 1$. The Orlicz norm of $f$ in $\psi_p$ is defined by
\bbeq\label{2Orlicz}
\|f\|_{\psi_p} :=\inf\{C>0;\ \int_S \psi_p\left(\frac{|f|}{C}\right)\leqslant 1\}.
\nneq

The following Bernstein's inequality due to Bolley-Villani and Gozlan-L\'eonard will be one of our main tools.

\bthm \label{thm_BV} {\bf ( Bolley-Villani \cite[Theorem 1(i)]{BolleyVillani05},  Gozlan-L\'eonard  \cite[Corollary 3]{GL07})} If $f$ is $\mu$-exponentially integrable, i.e. $ \|f\|_{\psi_1}<+\infty$, then the transport-entropy inequality \eqref{TE1} holds with
  \bbeq \label{3thmBern2bc}
  c_B=2 \|f\|_{\psi_1}^2 ,\qquad M= \|f\|_{\psi_1}.
  \nneq
\nthm

Let us recall some references and works on the transport-entropy inequalities and Bernstein's inequalities.

\brmk{\rm
The transport-entropy inequalities were introduced by Marton \cite{Mar96a, Mar96b} and Talagrand
\cite{Tal96a}, and turn out to be a powerful tool for quantitative study of measure concentration even in the dependent case, as well illustrated by Ledoux \cite{Ledoux99, Ledoux01}. The reader is also referred to \cite{DGW04} and the survey \cite{GL10} for history and references.
}\nrmk

\brmk{\rm
For the sharp Bernstein concentration inequalities for continuous time symmetric Markov processes, see Gao {\it et al.} \cite{GGW14}; for the (not-sharp) Bernstein concentration inequalities for discrete time Markov chains, see Wang and the first named author \cite{WangNYWu20}.  For sharp Bennett or Hoeffding's  concentration inequalities for martingales, see de la Pena {\it et al.} \cite{dPenaLaiShao09}, Rio \cite{Rio00} and the references therein.
}\nrmk

\brmk{\rm
Bernstein's inequality does not give an estimate of good order for large deviations.  For finer large deviation estimates the reader is referred to  Fan-Grama-Liu \cite{FGLiu14SC} and the references therein.
}\nrmk

Now we are ready to present the main results of this paper on $p_-(n,\vep)$.

\subsection{Bernstein's concentration inequality under Gaussian or exponential integrability.}

Our first result refines the estimate of Bernstein's constant $c_B$ in Theorem \ref{thm_BV}.

\bthm\label{thm_Ma}
\begin{enumerate}
  \item If $f$  is Gaussian integrable:
  $$\exists \delta>0:\ \ee e^{\delta f(Z)^2}=\int_S e^{\delta f^2(z)}\mu(dz)<+\infty
  $$
  then for any $\vep\in (0,\delta)$, \eqref{Bern-c} holds with
  \bbeq \label{thm_M1}
  c_B=\var_\mu(f)+\frac{1}{3} L(\vep),\qquad  M= \sqrt{\frac{2}{3\vep}}, 
  \nneq
  where
  \bbeq\label{thm_M2}
  L(\vep)=\frac{1}{\vep} \log \int_S e^{\vep(\tilde{f}^2-\mu(\tilde{f}^2))} d\mu, \qquad \tilde{f}:=f-\mu(f)
  \nneq
  satisfies for all $0<\vep<
  \frac{1}{\|\tilde{f}^2-\mu(\tilde{f}^2)\|_{\psi_1}}$,
  \bbeq\label{thm_M3}
  L(\vep)\leqslant \frac{\vep \|\tilde{f}^2-\mu(\tilde{f}^2)\|^2_{\psi_1} }{1- \vep\cdot\|\tilde{f}^2-\mu(\tilde{f}^2)\|_{\psi_1} }.
  \nneq
  Moreover for all $n\geqslant 1, x>0$ such that $0<x \leqslant \frac{n}{2}$, we have
  \bbeq\label{thm_M3b}
   \pp\left(L_n(f)-\mu(f)>\sqrt{2 \var_\mu(f) \frac{x}{n}} + 2\sqrt{\frac{\sqrt{2}\|\tilde{f}^2-\mu(\tilde{f}^2)\|_{\psi_1} }{3}} \left(\frac{x}{n} \right)^{3/4} \right)\le e^{-x}.
\nneq
  \item If $f\in L^2(S,\mu)$ and the positive part $f^+(x)=\max\{f(x),0\}$ is (only) exponentially integrable, i.e.
  $$
  \exists \delta>0:\ \ee e^{\delta f^+(Z)}=\int_S e^{\delta f^+(z)}\mu(dz)<+\infty,
  $$
  then for any $L>0$, setting $\vep(L):= \|f-f\wedge L\|_{\psi_1}$,  the Bernstein's concentration inequality \eqref{Bern-c} holds with
  \bbeq\label{thm_M4}
  \begin{cases}
  c_B&=\var_\mu(f)+ 2 \vep(L) \sqrt{2 \var_\mu(f)} + 2 \vep(L)^2\\
  &\leqslant (1+\vep(L)) \var_\mu(f)+  2(\vep(L) + \vep^2(L) )\\
   M&=\frac{L}{3} + \vep(L).
  \end{cases}
  \nneq
\end{enumerate}
\nthm

Its proof is postponed in Section 4.

\brmk {\rm
\begin{enumerate}
  \item Under the Gaussian integrability condition, the constants $c_B, M$ in \eqref{thm_M1} and \eqref{thm_M3b} are quite explicit, especially $c_B$ is sharp. Moreover \eqref{thm_M3b} gives a deviation inequality with  variance $\var_\mu(f)$ as the exact constant, which is  sharp in moderate deviations.

  \item Notice that $\vep(L)=\|f-f\wedge L\|_{\psi_1}\to 0$ as $L\to+\infty$, the concentration constant $c_B$ given by \eqref{thm_M4} becomes sharp in moderate deviations, too. A point in Theorem \ref{thm_Ma}(2) worth mentioning is that we do not require the exponential integrability of $f$ for the one-sided Bernstein's concentration inequality, which is quite natural as seen from Theorem \ref{thmA}.

  \item Our approach for this theorem will be based on large deviations and on weighted Csiszár-Kullback-Pinsker's (CKP in short) transport-entropy inequality, developed in \cite{BolleyVillani05, DGW04, GL07}.
  \item For comparison with the previously known Bernstein type inequality of Adamczak \cite{Ada08}  in the unbounded case, see Remark \ref{adamczak}. 
\end{enumerate}

}
\nrmk

\subsection{Dimension-free lower bound of the minimal risk: sharp Bernstein's inequalities}

We now apply the previous Bernstein's concentration inequalities to dimension-free estimates of $p_-(n,\vep)$, i.e. an estimate for the lower bound of the theoretical minimal risk $\inf_{\theta\in \Theta} R(\theta)$. We begin with the upper boundedness or the uniform Gaussian integrability case.

\bthm\label{thmM21} Assume {\bf (H1)} and
\bbeq\label{thmM21c}
\M:=\argmin_{\theta\in \Theta} R(\theta)\neq \emptyset
\nneq
i.e. the optimal parameters exist.
\begin{enumerate}
  \item {\bf (Under the local upper-boundedness)} If there is some constant $b>0$ such that for all $\theta_0\in \M$,
  $$
  Q(z,\theta_0) \leqslant b, \qquad \mu-a.s.
  $$
  then for any $x>0, \ n\geqslant 1$,
  \bbeq\label{thm29a}
  \pp\left(\inf_{\theta\in \Theta} R(\theta) \geqslant \inf_{\theta\in \Theta} R_{E,n}(\theta) - \vep(n,x) \right)\geqslant 1-e^{-x}
  \nneq
  or equivalently,
  \bbeq\label{thmM21a}
 p_-(n,\vep(n, x))=\pp\left(\inf_{\theta\in \Theta} R(\theta) < \inf_{\theta\in \Theta} R_{E,n}(\theta) - \vep(n,x) \right)\leqslant e^{-x},
  \nneq
  where
  \bbeq\label{thm29b}
  \vep(n,x):= \inf_{\theta_0\in \M} \|Q_{\theta_0}\|_{L^2(\mu)} \sqrt{\frac{2x}{n}} + \frac{b}{3} \cdot \frac{x}{n}.
  \nneq
  \item {\bf (Under the local Gaussian integrability)} If $\{Q_{\theta_0};\ \theta_0\in\M\}$ is of uniform Gaussian integrability:
  \bbeq\label{thm29c}
  \exists \lambda_0>0:\  \sup_{\theta_0\in \M}\ee e^{\lambda_0 Q(Z,\theta_0)^2}= \sup_{\theta_0\in \M}\int_S e^{\lambda_0 Q(z, \theta_0)^2}\mu(dz)<+\infty
  \nneq
  then for any $x>0, \ n\geqslant 1$, and
  $$0<\delta<
\min\left\{
\lambda_0,\,
\frac{1}{
2\sup_{\theta_0\in \M}
\|\widetilde Q_{\theta_0}^2-\mu(\widetilde Q_{\theta_0}^2)\|_{\psi_1}
}
\right\},$$
  we have
  \bbeq\label{thmM21b}
  \aligned
  &\pp\left(\inf_{\theta\in \Theta} R(\theta) \geqslant \inf_{\theta\in \Theta} R_{E,n}(\theta) - \vep(n,x)\right)\geqslant 1-e^{-x} \\
  \endaligned
  \nneq
  where $\sigma^2(\theta_0)= \var(Q_{\theta_0})$ and
  \bbeq\label{thmM21b1}
  \aligned
  \vep(n,x):=&
  \left(\frac{2\left(\inf_{\theta_0 \in \M}\sigma^2(\theta_0)+ \frac{2\delta}{3} \sup_{\theta_0 \in \M} \|\tilde{Q}_{\theta_0}^2-\mu(\tilde{Q}_{\theta_0}^2)\|^2_{\psi_1}\right) x}{n}\right)^{1/2}\\
   &\quad + \sqrt{\frac{2}{3\delta}} \frac{x}{n}.
  \endaligned
  \nneq

  Moreover for all $n\geqslant 1, x>0$ such that $0<x \leqslant \frac{n}{2}$, we have
  \bbeq\label{thm_M21e}
   \pp\left(\inf_{\theta\in \Theta} R(\theta) \geqslant \inf_{\theta\in \Theta} R_{E,n}(\theta) - \vep(n,x) \right)\geqslant 1-e^{-x},
\nneq
where
\bbeq\label{thm_M21f}
\vep(n,x):= \inf_{\theta_0\in\M} \left(\sigma(\theta_0)\sqrt{ \frac{2x}{n}} + 2\sqrt{\frac{\sqrt{2} \|\tilde{Q}_{\theta_0}^2-\mu(\tilde{Q}_{\theta_0}^2)\|_{\psi_1} }{3}} \left(\frac{x}{n} \right)^{3/4}\right).
\nneq

\end{enumerate}

\nthm

\brmk\label{2R28}{\rm
\begin{enumerate}
  \item Part (1) above is an easy consequence of the classical result: Theorem \ref{thmA}. In fact
let $\theta_0 \in \M$ be an arbitrary minimizer of $R(\theta)$, i.e., $R(\theta_0)\leqslant R(\theta), \forall \theta\in \Theta$. In that case
$$
\aligned
p_-(n,\vep) &= \pp\left( \inf_{\theta\in \Theta} R_{E,n}(\theta)> \inf_{\theta\in \Theta} R(\theta) + \vep\right)\\
&\leqslant \pp\left( R_{E,n}(\theta_0)> R(\theta_0) + \vep\right)
\endaligned
$$
which reduces to the estimate of a single observable $Q(z,\theta_0)$ for which Theorem \ref{thmA} is applicable and gives us part (1).
  \item
Part (2) with  a singleton $\Theta$ generalizes Theorem \ref{thmA}. It gives a sharp Bernstein concentration inequality without the (upper) boundedness assumption of $Q$, which is new to the best of our knowledge.

  \item Under the Gaussian integrability, using the transport-entropy inequality approach (as in \cite{BolleyVillani05}, \cite{DGW04}, \cite{GL07}), we can derive a Hoeffding's concentration inequality. But unfortunately the constant $c_B$ (with $M=0$) does not involve $\sigma^2(\theta_0)$, which is weaker (for small $x$) than the results in Theorem \ref{thmM21} in the minimal risk problem.
\end{enumerate}
}
\nrmk

\bprf[Proof of Theorem \ref{thmM21}]
Part (1) has been established in Remark \ref{2R28}.

For Part (2), by {\bf (H1)},
$\inf_{\theta\in\Theta} R_{E,n}(\theta)$, $n\geqslant 1$ are random variables.
Let $\theta_0 \in \M$ be an arbitrary minimizer of $R(\theta)$ over $\Theta$. We have
$$
p_-(n, \vep):=\pp\left( \inf_{\theta\in \Theta} R_{E,n}(\theta)> \inf_{\theta\in \Theta} R(\theta)+\vep  \right)\leqslant \pp\left( R_{E,n}(\theta_0)> R(\theta_0)  +\vep \right).
$$
For this fixed $\theta_0\in\M$, for $\delta\in (0,\lambda_0)$ let 
\bbeq\label{thm_M21c}
L(\theta_0, \delta):=\frac{1}{\delta}\log\int e^{\delta(\tilde Q_{\theta_0}^{\,2}-\mu(\tilde Q_{\theta_0}^{\,2}))}\,d\mu,
\qquad
\tilde Q_{\theta_0}:=Q(z,\theta_0)-R(\theta_0).
\nneq
Then for all
$$
0<\delta<\frac{1}{2\sup_{\theta_0 \in \M}\|\tilde Q_{\theta_0}^{\,2}-\mu(\tilde Q_{\theta_0}^{\,2})\|_{\psi_1}},
$$
by \eqref{thm_M3} we have 
\bbeq\label{thm_M21d}
L(\theta_0, \delta)
\leqslant
\frac{\delta\|\tilde Q_{\theta_0}^{\,2}-\mu(\tilde Q_{\theta_0}^{\,2})\|_{\psi_1}^{\,2}}
{1-\delta\|\tilde Q_{\theta_0}^{\,2}-\mu(\tilde Q_{\theta_0}^{\,2})\|_{\psi_1}}
\leqslant
2\delta\|\tilde Q_{\theta_0}^{\,2}-\mu(\tilde Q_{\theta_0}^{\,2})\|_{\psi_1}^{\,2}.
\nneq
Applying Theorem \ref{thm_Ma} to the r.h.s., we get the two concentration inequalities
\eqref{thmM21b} and \eqref{thm_M21e}.
\nprf

Since the Bernstein's inequality is equivalent to the exponential integrability, we now generalize Theorem \ref{thmM21} to the exponential integrability case.

\bthm\label{thmM22} Assume {\bf (H1)} and $\M \neq \emptyset$.
If the positive parts 
$$Q^+(z,\theta_0)=\max\{Q(z,\theta_0),0\},\qquad \theta_0\in\M$$ are uniformly exponentially integrable, i.e.
   \bbeq\label{thm_M22a}
  \exists \lambda_0>0:\ \sup_{\theta_0\in\M}\ee e^{\lambda_0 Q^+(Z,\theta_0)}=\sup_{\theta_0\in\M}\int_S e^{\lambda_0 Q^+(z,\theta_0)}\mu(dz)<+\infty,
  \nneq
  then for any $L>0$, $x>0$ and $n\geqslant1$, setting
  $$\vep(\theta, L):= \| Q_{\theta}-  Q_{\theta}\wedge L\|_{\psi_1},\qquad \vep(L):= \sup_{\theta_0\in\M} \vep(\theta_0, L)$$
  the following Bernstein's concentration inequality  holds

  \bbeq\label{thm_M22b}
  \aligned
\pp\left(\inf_{\theta\in \Theta} R(\theta) \geqslant \inf_{\theta\in \Theta} R_{E,n}(\theta) - \vep(n,x) \right)\geqslant 1-e^{-x}
\endaligned
  \nneq
where
  \bbeq\label{thm_M22c}
  \begin{cases}
  &\vep(n,x):= \sqrt{\frac{2 \inf_{\theta_0\in \M} c_B(\theta_0) x}{n}} + M \frac{x}{n}\\
  &c_B(\theta_0)= (1+\vep(L)) \var_{\mu}(Q_{\theta_0}^2) +  2(\vep(L) + \vep(L)^2 )\\
   &M=\frac{L}{3} + \vep(L).
  \end{cases}
  \nneq

\nthm

\bprf[Proof of Theorem \ref{thmM22}] As in the proof of Theorem \ref{thmM21}, we have for any $\theta_0\in\M$,
 $$
p_-(n, \vep):=\pp\left( \inf_{\theta\in \Theta} R_{E,n}(\theta)> \inf_{\theta\in \Theta} R(\theta)+\vep  \right)\leqslant \pp\left( R_{E,n}(\theta_0)> R(\theta_0)  +\vep \right).
$$
By Theorem \ref{thm_Ma} (2), we can obtain the desired Bernstein's concentration inequality \eqref{thm_M22c} by optimizing over $\theta_0\in \M$.

\nprf

\brmk{\rm The main difference from Bousquet's inequality and Klein-Rio's inequality (usually essential for sharp estimates of $p_+(n,\vep)$) is that only the minimal variance $\inf_{\theta_0\in \M}\var_\mu(Q_{\theta_0})$  in the localized minimizer set $\M$ is used here, instead of the maximal variance $\sigma^2(\Theta)$ over $\Theta$, in the leading term of $\vep(n,x)$. 
}\nrmk

\subsection{We can quickly verify the non-efficiency of a learning machine}

To verify if a learning machine does not work efficiently for approximating a given function $y=f(x)$, fix the risk function $Q(z,\theta)=|y-f(x,\theta)|^2$ or $Q(z,\theta)=|y-f(x,\theta)|$. Usually $Q(z,\theta)=|y-f(x,\theta)|$ satisfies the Gaussian integrability condition, $Q(z,\theta)=|y-f(x,\theta)|^2$ satisfies the exponential integrability condition. Assume one of them.

Given a small error probability level $\alpha\in (0, 0.5)$ and a tolerance error $\vep_0>0$, we take a sample of
$$
Y_i=f(X_i), \ 1\leqslant i\leqslant n
$$
where the sample size $n$ satisfies
$$
\vep\left(n, \log\frac{1}{\alpha}\right)\leqslant \vep_0.
$$
Here $\vep(n,x)$ are given in Theorem \ref{thmM21} or \ref{thmM22}. Roughly
$$n \gtrsim \frac{\inf_{\theta_0\in \M}\sigma^2(\theta_0)}{\vep_0^2} \log \frac{1}{\alpha},$$  which is dimension-free and slightly greater than the famous asymptotic  ''three-sigma" rule $n\geqslant 9\sigma^2/\vep_0^2$, corresponding approximately to a $99.7\%$ confidence level under the central limit
theorem. Then by Theorem \ref{thmM21} or \ref{thmM22}, the minimal risk $\inf_{\theta\in\Theta} R(\theta)$ of the learning machine satisfies
$$
\inf_{\theta\in\Theta} R(\theta)\geqslant \inf_{\theta\in\Theta} R_{E,n}(\theta)-\vep_0
$$
with confidence level $\geqslant 1-\alpha$.

\bigskip
{\bf Conclusion:} {\it if $\inf_{\theta\in\Theta} R_{E,n}(\theta)$ is not small with a sample size $n\gtrsim\frac{\inf_{\theta_0\in \M}\sigma^2(\theta_0)}{\vep_0^2} \log \frac{1}{ \alpha}$  (dimension-free), then the learning machine is most probably not efficient for the approximation of a deterministic function $y=f(x)$.}

\section{Upper bound of the minimal risk in terms of $d_{\psi_1}$-metric entropies}

As recalled in the Introduction, if $\HH=\{Q(z,\theta); \theta\in \Theta\}$ is bounded in $L^\infty(\mu)$, one should work with a sample size
$$n\succeq \frac{1}{\vep_0^2} \vc(\HH)$$
(i.e. much greater than $\frac{1}{\vep_0^2} \vc(\HH)$),
for verifying the efficiency of a learning machine (within precision $\vep_0$), if we use the VC dimension. Recent progress consists in replacing $\vc(\HH)$ by distribution dependent or even data-driven metric entropy, see \cite{Kol11, BLM13, Wainwright19} and the references therein.

\subsection{Packing number and metric entropy}

For the reader's convenience (also for stating our results), we recall the following notions.

\bdef {\bf (covering number, packing number and metric entropy)}
Fix a semi-metric $d$ on $\Theta$ (i.e. ''semi" means that $d(\theta_1,\theta_2)=0$ may hold for two different points of $\Theta$), $\vep>0$.

\begin{enumerate}
  \item
 An $\vep$-net of $\Theta$ is a subset $\mathcal{N}$ of $\Theta$ such that
$$
\forall \theta\in \Theta, \exists \theta_k\in \mathcal{N} \text{ such that } d(\theta, \theta_k)\leqslant \vep.
$$
The smallest possible cardinality $N(\Theta, d, \vep)$ of $\vep$-nets of $\Theta$  is also the smallest number of
closed balls of $d$-radius $\vep$, whose union covers $\Theta$. $N(\Theta, d, \vep)$ is called the {\bf covering number} of $\Theta$.

  \item A subset $D$ of $\Theta$ is called ''{\it $\vep$-separated}", if the distance between each pair of points in $D$ is strictly larger than $\vep$. The maximal possible
  cardinality of $\vep$-separated subsets of $\Theta$ is called the {\bf packing number} of $\Theta$, denoted by ${\mathcal P}(\Theta, d, \vep)$.

  \item
Their logarithms
$$\log N(\Theta, d, \vep), \qquad \log {\mathcal P}(\Theta, d, \vep)$$
are called {\bf metric entropies}.
\end{enumerate}
\ndef
It is well known that (see e.g. \cite[p. 98]{vdVW96} or \cite[Lemma 4.2.8]{Vershynin20})
\bbeq\label{PN224}
N(\Theta, d, \vep)\leqslant {\mathcal P}(\Theta, d, \vep)\leqslant N(\Theta, d, \vep/2).
\nneq
And $(\Theta, d)$ is totally bounded if and only if for any $\vep>0$, both the covering number $N(\Theta, d, \vep)$ and the packing number ${\mathcal P}(\Theta, d, \vep)$ are finite.

On $\Theta$ we consider the following two semi-metrics 
\bbeq\label{d2}
\aligned
d_2(\theta_1,\theta_2) & = \|Q(\cdot,\theta_1) - Q(\cdot,\theta_2)\|_2\\
d_{\psi_1} (\theta_1,\theta_2) &= \| Q(\cdot,\theta_1) - Q(\cdot,\theta_2) \|_{\psi_1}\\
\endaligned
\nneq

By Taylor's expansion of $e^x-1$ at $x=0$, we have immediately
\bbeq\label{d2-psi1}
d_2(\theta_1,\theta_2) \leqslant \sqrt{2} d_{\psi_1} (\theta_1,\theta_2),\qquad \forall \theta_1,\theta_2\in\Theta.
\nneq

\subsection{$L_{\psi_1}$-estimates of empirical risk in terms of metric entropy}
We will directly study the concentration of the normalized empirical process
\bbeq\label{2W}
W_n(\theta) = \frac{1}{\sqrt{n}} \sum_{k=1}^{n} (Q(Z_k,\theta)-R(\theta))= \sqrt{n}(R_{E,n}(\theta)- R(\theta))
\nneq
as in the studies on the central limit theorem of empirical processes. We introduce 

\medskip
{\bf (H2)} {\it  $\HH=\{Q_\theta=Q(\cdot, \theta);\ \theta\in\Theta\}$ is bounded in $L_{\psi_1}(\mu)$, i.e.
\bbeq\label{thmM23a1}
\sup_{\theta\in \Theta} \|Q_\theta\|_{\psi_1}<+\infty.
\nneq
}
\medskip

Let
$${\rm Diam}(\Theta, d)={\rm D}(\Theta, d):= \sup\{d(\theta_1,\theta_2);\ \theta_1,\theta_2\in \Theta\}$$
be the diameter of $\Theta$ in the pseudo metric $d$. It is necessarily finite under \eqref{thmM23a1} for $d$ being one of two metrics given in \eqref{d2}.  

We begin with

\blem\label{lem31}
{\bf (essentially contained in \cite[Lemma 13.1]{BLM13})}
Assume {\bf (H1)} and {\bf (H2)}. Then the following estimate holds.

For any $n\geqslant 1$,
\bbeq\label{lem31a}
\ee \sup_{\theta\in \Theta} W_n(\theta)
\leqslant
12 \sqrt{2}
\int_{0}^{{\rm D}(\Theta, d_{\psi_1})/2}
\sqrt{\log \mathcal{P}(\Theta, d_{\psi_1},\vep)}\,d\vep
+
\frac{12}{\sqrt{n}}
\int_{0}^{{\rm D}(\Theta, d_{\psi_1})/2}
\log \mathcal{P}(\Theta, d_{\psi_1},\vep)\,d\vep.
\nneq
\nlem

\bprf[Proof of Lemma \ref{lem31}]
Below we verify the condition in \cite[Lemma 13.1]{BLM13}.

By Theorem \ref{thm_BV} applied to $Q_{\theta_1}-Q_{\theta_2}$, we have
$$
\log \ee e^{\lambda (\widetilde{Q}_{\theta_1}-\widetilde{Q}_{\theta_2})}
\leqslant
\frac{
2d_{\psi_1}(\theta_1,\theta_2)^2\lambda^2
}{
2\left(1-\lambda d_{\psi_1}(\theta_1,\theta_2)\right)
},
\qquad
\lambda\in\left(0,1/d_{\psi_1}(\theta_1,\theta_2)\right).
$$
This implies, by \eqref{Bern-a2} in Theorem \ref{thm_GL}, that
$$
\log \ee e^{\lambda (W_n(\theta_1)-W_n(\theta_2))}
\leqslant
\frac{
2d_{\psi_1}(\theta_1,\theta_2)^2\lambda^2
}{
2\left(1-\lambda d_{\psi_1}(\theta_1,\theta_2)/\sqrt n\right)
},
\qquad
\lambda\in\left(0,\sqrt n/d_{\psi_1}(\theta_1,\theta_2)\right).
$$
The condition in \cite[Lemma 13.1]{BLM13} is verified, and \eqref{lem31a} follows.

\nprf

The following result may be interesting in the theory of empirical processes.

\bthm\label{thmM23} Assume {\bf (H1)} and {\bf (H2)} (i.e. $\HH$ is bounded in $L_{\psi_1}(\mu)$).

Then, for the normalized empirical process $(W_n(\theta), \theta\in\Theta)$ given in \eqref{2W}, the following estimates hold for any sample size  $n\geqslant 2$.
\begin{enumerate}
\item For any $\theta_0\in \Theta$, if $\pm \tilde{Q}_{\theta_0}$ satisfy the Bernstein inequality \eqref{Bern-a} with constants $(c_B(\theta_0), M(\theta_0))$, then
\bbeq\label{thmM23i}
\aligned
\|W_n(\theta_0)\|_{\psi_1} \leqslant  \max\left\{2 \sqrt{\frac{c_B(\theta_0)}{\log 2}},  \frac{4 M(\theta_0) }{\sqrt{n}}\right\}.
\endaligned
\nneq
In particular, for $c_B(\theta_0)=2\|Q_{\theta_0}\|_{\psi_1}^2, \ M(\theta_0)= \|Q_{\theta_0}\|_{\psi_1}$ (by Theorem \ref{thm_BV}),
\bbeq\label{thmM23i2}
\aligned
\|W_n(\theta_0)\|_{\psi_1} \leqslant  2 \sqrt{\frac{2}{\log 2}} \cdot \|Q_{\theta_0}\|_{\psi_1}.
\endaligned
\nneq

\item
\bbeq\label{thmM23a}
\aligned
\|\sup_{\theta\in \Theta} |W_n(\theta)|\|_{\psi_1}
&\leqslant   2 \sqrt{\frac{2}{\log 2}}\cdot \inf_{\theta_0\in \Theta} \|Q_{\theta_0}\|_{\psi_1}
+\frac{8\sqrt{6}}{\log 2} \int_{0}^{{\rm D}(\Theta, d_{\psi_1} )} \sqrt{ \log (1+{\mathcal P}(\Theta, d_{\psi_1}, \vep))} d\vep\\
& + \frac{48}{(\log 2)\sqrt{n}} \int_{0}^{{\rm D}(\Theta, d_{\psi_1} )} \log(1+{\mathcal P}(\Theta, d_{\psi_1},\vep)) d\vep.
\endaligned
\nneq

\item
\bbeq\label{thmM23b2}
\aligned
\ee \sup_{\theta\in \Theta} |W_n(\theta)|
&\leqslant \inf_{\theta_0\in \Theta} \sigma(\theta_0) + \frac{8\sqrt{6}}{\log 2} \int_{0}^{{\rm D}(\Theta, d_{\psi_1} )} \sqrt{ \log (1+{\mathcal P}(\Theta, d_{\psi_1}, \vep))} d\vep\\
& + \frac{48}{(\log 2)\sqrt{n}} \int_{0}^{{\rm D}(\Theta, d_{\psi_1} )} \log(1+{\mathcal P}(\Theta, d_{\psi_1},\vep)) d\vep.
\endaligned
\nneq

\item For any $\delta, \eta>0$,
\bbeq\label{thmM23c}
\aligned
&\|\sup_{\theta, \theta'\in \Theta: d_{\psi_1}(\theta,\theta')\leqslant \delta} |W_n(\theta)- W_n(\theta')|\|_{\psi_1}\\
&\leqslant \frac{2\sqrt{6}}{\log 2} \left( 16\int_{0}^{\eta} \sqrt{ \log (1+{\mathcal P}(\Theta, d_{\psi_1}, \vep))} d\vep + \delta \sqrt{\log (1+{\mathcal P}(\Theta, d_{\psi_1}, \eta)^2)} \right)\\
&\quad +  \frac{12}{(\log 2)\sqrt{n}}\left( 16\int_{0}^{\eta} \log(1+{\mathcal P}(\Theta, d_{\psi_1}, \vep)) d\vep + \delta \log(1+{\mathcal P}(\Theta, d_{\psi_1}, \eta)^2)\right)
\endaligned
\nneq
\end{enumerate}
\nthm

Its proof is postponed in Section 5.

Let us make some comments on this theorem.

\brmk{\rm
\begin{enumerate}
\item For a general centered stochastic process $(X(\theta); \theta\in \Theta)$ satisfying the ''sub-Gaussian" condition w.r.t. a semi-metric $d$ on $\Theta$,
      \bbeq\label{sub-Gauss}
      \pp(|X(\theta_1) -X(\theta_2)|> x) \leqslant 2 e^{-x^2/(2 d^2(\theta_1, \theta_2))},
      \nneq
(here the constants $2$, $1/2$ have no importance), then we have the following well-known Dudley's inequality (\cite{Vershynin20})
$$
\ee \sup_{\theta\in \Theta} X(\theta) \leqslant K \int_{0}^{{\rm D}(\Theta, d)} \sqrt{ \log (1+{\mathcal P}(\Theta, d, \vep))} d\vep
$$
for some absolute constant $K$, and also \eqref{thmM23c} for $X(\theta)$ (instead of $W_n(\theta)$),  without the extra term
$$ \frac{K}{\sqrt{n}} \int_{0}^{{\rm D}(\Theta, d)} \log(1+{\mathcal P}(\Theta, d,\vep)) d\vep$$ (\cite[Theorem 2.2.4 and Corollary 2.2.8]{vdVW96}).

For $n=1$, $W_n(\theta)=\tilde{Q}_\theta$ satisfies the ''sub-Gaussian" condition \eqref{sub-Gauss} w.r.t. the semi-metric
$$d_{\psi_2}(\theta_1, \theta_2):=\|\tilde Q_{\theta_1} - \tilde Q_{\theta_2} \|_{\psi_2}.$$
In fact by Chebyshev's inequality and the fact that $\ee \exp\left(X^2/\|X\|^2_{\psi_2}\right)\leqslant 2$, we have
$$
\pp\left(|X|>x\right) \leqslant 2 \exp\left(- x^2/\|X\|^2_{\psi_2}\right), \ \forall x>0.
$$
Applying it to $X=\tilde{Q}_{\theta_1}- \tilde{Q}_{\theta_2}$, we obtain the sub-Gaussian condition \eqref{sub-Gauss} with $d(\theta_1, \theta_2)=\|\tilde{Q}_{\theta_1}- \tilde{Q}_{\theta_2}\|_{\psi_2} $.

But unfortunately it is very difficult to prove that $W_n(\theta)$ satisfies the sub-Gaussian condition w.r.t. a reasonable metric $d$ except for the Rademacher sequence (guaranteed by Hoeffding's inequality), for general $n\geqslant 2$.
Here ''reasonable" means that the involved metric $d$ is independent of the sample size $n$, and the packing number ${\mathcal P}(\Theta, d,\vep)$ should be computable and controllable. That is why the symmetrization technique based on Rademacher sequence is important in the theory of empirical processes (see \cite{vdVW96} and also the later works \cite{BM02, BBM05, Kol06}).

Theorem \ref{thmM23} is new for general $n\geqslant 2$ in the exponentially integrable case.  Its proof will be based again on the classical {\bf chaining} technique, with the help of Bernstein's inequalities in Theorem \ref{thm_BV} and Theorem \ref{lem41}, replacing the sub-Gaussian condition  and the classic symmetrization technique. This approach was developed by Boucheron-Lugosi-Massart \cite[Corollary 2.6, Lemma 13.1]{BLM13} for obtaining the estimate  of the expectation of supremum of empirical processes.

\item The first term $ K \int_{0}^{\eta} \sqrt{ \log (1+{\mathcal P}(\Theta, d_{\psi_1}, \vep))} d\vep$ in \eqref{thmM23a} and \eqref{thmM23c}
comes from the Gaussian regularization of  $W_n(\theta)$, the extra term $ \frac{K}{\sqrt{n}} \int_{0}^{{\rm D}(\Theta,d_{\psi_1})} \log(1+{\mathcal P}(\Theta, d_{\psi_1},\vep)) d\vep$ can be interpreted as the discrepancy from the Gaussian  regularization.

  \item The bound \eqref{thmM23c} in part (4) with $\delta=\eta={\rm Diam}(\Theta, d_{\psi_1})$ gives \eqref{thmM23a} with an extra constant term ${\rm D}(\Theta,d_{\psi_1})\log 2$. We state \eqref{thmM23a} separately to make the proof clearer.
      
  \item Typically, classical results of the form stated in Theorem \ref{thmM23} contain non‑explicit absolute constants. In this work, we render all involved constants explicit.

\end{enumerate}
}
\nrmk

\subsection{Estimates of empirical risk in terms of the box-dimension} 

Recall the {\it Minkowski or box dimension} of $(\Theta, d)$ (\cite{Falconer}):
\bbeq\label{box_dim}
\dim_B(\Theta, d) = \limsup_{\vep\to 0+} \frac{\log (1+N(\Theta, d, \vep))}{-\log \vep}= \limsup_{\vep\to 0+} \frac{\log (1+ {\mathcal P}(\Theta, d, \vep))}{-\log \vep}.
\nneq

\bcor\label{cor219}
In the framework of Theorem \ref{thmM23}, assume that $d_B:=\dim_B(\Theta, d_{\psi_1})<+\infty$. Then there is some constant $C>0$ such that for all $n\geqslant 2$:
\bbeq\label{thmM23e}
\ee \sup_{\theta\in \Theta} |R_{E,n}(\theta)-R(\theta)| \leqslant \frac{1}{\sqrt{n}}\left(\inf_{\theta_0\in \Theta}\sigma(\theta_0) + \frac{8\sqrt{6}}{\log 2}\sqrt{CD(1+d_B)}\right) + 
 \frac{ 48 CD(1+d_B)} {(\log 2)n}.
\nneq
and for any $x>0$, $n\geqslant 2$,

\bbeq\label{thmM23f}
\aligned
\pp&\left( \sup_{\theta\in \Theta} |R_{E,n}(\theta)-R(\theta)| >
\right.\\
&\quad\left. x \left[ 2\sqrt{\frac{2}{n\log 2}} \inf_{\theta_0\in \Theta}  \|Q_{\theta_0}\|_{\psi_1}+  \left(\frac{8\sqrt{6}}{\log 2} \sqrt{\frac{CD(1+ d_B)}{n}} + \frac{ 48 CD(1+d_B)} {(\log 2)n}\right)\right]
\right)\\
& \leqslant 2e^{-x}.
\endaligned
\nneq
Here the constant $C$ is determined by \eqref{cor219b} below and $D:={\rm D}(\Theta, d_{\psi_1})$ is the diameter. 
\ncor
The uniform estimate \eqref{thmM23f} implies that if
\bbeq\label{thmM23f2}
n \gtrsim \frac{1+\dim_B(\Theta, d_{\psi_1})}{\vep_0^2} 
\nneq
(instead of ${\rm vc}(\mathcal{H})$ recalled in Introduction), we have a non-asymptotic but non-explicit (because of the presence of the constant $C$) estimate of the error probabilities $p_{\pm}(n,\vep_0)$, which are small.

The above corollary generalizes Theorem \ref{3thm_VCbias} in two aspects: the VC dimension is replaced by the Minkowski  dimension; and the boundedness assumption is replaced by
the exponential ($L_{\psi_1}$-)integrability condition.

\bprf Write $d_B=\dim_B(\Theta, d_{\psi_1})$ for simplicity. 
To begin with, by the definition of the box dimension,
$$
\log (1+ {\mathcal P}(\Theta, d_{\psi_1}, \vep)) \leqslant C_1 (d_B+1) \log\frac{1}{\vep}, \ \forall \vep\in (0,1/e),
$$
for some positive constant $C_1$. Therefore, for some constant $C>0$ (depending on $C_1$ and $D$), we have  
\bbeq\label{cor219b}
\aligned
&\int_{0}^{D} \log (1+{\mathcal P}(\Theta, d_{\psi_1}, \vep)) d\vep \leqslant CD (1+d_B)\\
&\int_{0}^{D} \sqrt{ \log (1+{\mathcal P}(\Theta, d_{\psi_1}, \vep))} d\vep \leqslant  \sqrt{CD(1+d_B)}\\
\endaligned
\nneq
(the latter follows from the first by Cauchy-Schwarz inequality, after enlarging $C$ if necessary). 
Noting that $$R_{E,n}(\theta)-R(\theta)=\frac{1}{\sqrt{n}}W_n(\theta),$$ by \eqref{thmM23b2} we have
$$
\aligned
\ee \sup_{\theta\in \Theta} |R_{E,n}(\theta) - R(\theta)|
&\leqslant \frac{\inf_{\theta_0\in \Theta}\sigma(\theta_0)} {\sqrt{n}}+ \frac{8\sqrt{6}}{(\log 2)\sqrt{n}} \int_{0}^{D} \sqrt{ \log (1+{\mathcal P}(\Theta, d_{\psi_1}, \vep))} d\vep\\
& + \frac{48}{(\log 2)n} \int_{0}^{D} \log(1+{\mathcal P}(\Theta, d_{\psi_1},\vep)) d\vep\\
&\leqslant   \frac{1}{\sqrt{n}}\left(\inf_{\theta_0\in \Theta}\sigma(\theta_0) + \frac{8\sqrt{6}}{\log 2}\sqrt{CD(1+d_B)}\right) +  \frac{ 48 CD(1+d_B)} {(\log 2)n}.
\endaligned
$$
Therefore, we obtain \eqref{thmM23e}.

By Chebyshev's inequality and the definition of Orlicz norm  $\|\cdot\|_{\psi_1}$,  we have
$$
\aligned
\pp\left( \sup_{\theta\in \Theta} |R_{E,n}(\theta)-R(\theta)| > x\cdot \|\sup_{\theta\in \Theta} |R_{E,n}(\theta)-R(\theta)| \|_{\psi_1}\right)
 \leqslant 2e^{-x}  ,\ x>0.
\endaligned
$$
Since $\|\sup_{\theta\in \Theta} |R_{E,n}(\theta)-R(\theta)| \|_{\psi_1}= \frac{1}{\sqrt{n}} \|\sup_{\theta\in \Theta} |W_n(\theta)| \|_{\psi_1}$, applying \eqref{thmM23a} and \eqref{thmM23i2}
we obtain \eqref{thmM23f}.
\nprf

\brmk{\rm Assume that $\theta\to Q_\theta$ is Lipschitzian from $\Theta\subset \rr^N$ to $L_{\psi_1}$: $\exists L>0$ such that 
$$
\|Q_{\theta_1}-Q_{\theta_2}\|_{\psi_1} \leqslant L |\theta_1-\theta_2|, \ \forall \theta_1,\theta_2\in \Theta
$$
and $\Theta$ is bounded (that is not a real restriction, because in applications we can take $\Theta=\{\theta| R(\theta)\leqslant b\}$ for some great risk level $b$, which is usually bounded).

In that case we have $\mathcal{P}(\Theta, d_{\psi_1}, \vep) \leqslant \mathcal{P}(\Theta, |\cdot|, \vep/L)$ and then $d_B=\dim_B(\Theta, d_{\psi_1})\leqslant N$. Furthermore the sharp estimates of the packing number 
$\mathcal{P}(\Theta, |\cdot|, \vep)$ are classic results (for example $\mathcal{P}(B(0,R), |\cdot|, \vep) \leqslant (1+2R/\vep)^N$ for any $0<\vep<R$, for the ball $B(0,R)$ centered at $0$ of radius $R$).   
}\nrmk

\subsection{Talagrand’s concentration inequality for $p_+(n,\vep)$: upper bound of the minimal risk in the lower bounded case}

A natural question is to sharpen the exponential concentration inequality \eqref{thmM23f} into a much better Gaussian one for moderate $x=o(n)$, as in Bernstein's or Talagrand's concentration
inequality. This is the purpose of this subsection.

Let
$$
R_*=\inf_{\theta\in\Theta}R(\theta)
$$
be the theoretical true minimal risk, and set
$$
R^*=\sup_{\theta\in\Theta}R(\theta).
$$
Under {\bf (H2)}, both quantities are finite.

In this paragraph we assume that the risk functions are uniformly lower bounded:
\bbeq\label{thmM26a}
Q_\theta(z)=Q(z,\theta)\geqslant -L,\qquad z\in S,\ \theta\in\Theta,
\nneq
for some constant $L>0$. This is the case, for instance, when
$Q(z,\theta)=|y-f(x,\theta)|^\alpha$.

Our estimate of $p_+(n,\vep)$ is based on the following risk-level decomposition (say, peeling argument).
For any $\delta>0$, let
$$
M:=\min\left\{m\in\mathbb N:\ R_*+\sqrt{m+1}\delta\geqslant R^*\right\}.
$$
Define
for $0\leqslant k\leqslant M$,
\bbeq\label{3eq19}
\M_k
:=
\left\{
\theta\in\Theta:
R_*+\sqrt{k}\delta
\leqslant R(\theta)
\leqslant R_*+\sqrt{k+1}\delta
\right\}
\nneq
(hence $\M_0=\{\theta\in\Theta| R(\theta)\leqslant R_*+\delta\}$). 
Then
$$
\Theta=\bigcup_{k=0}^{M}\M_k \ \text{ and }\ \inf_{\theta\in\Theta}R_{E,n}(\theta)
=
\min_{0\leqslant k\leqslant M}
\inf_{\theta\in\M_k}R_{E,n}(\theta)
$$
where the convention
$
\inf \emptyset=+\infty$ is used. 
Hence
\bbeq\label{thmM26b}
\aligned
p_+(n,\vep)
&=
\pp\left(
R_*>\inf_{\theta\in\Theta}R_{E,n}(\theta)+\vep
\right)
\\
&\leqslant
\sum_{k=0}^{M}
\pp\left(
R_*>\inf_{\theta\in\M_k}R_{E,n}(\theta)+\vep
\right)
\\
&\leqslant
\pp\left(
R_*>\inf_{\theta\in\M_0}R_{E,n}(\theta)+\vep
\right)
\\
&\quad+
\sum_{k=1}^{M}
\pp\left(
\inf_{\theta\in\M_k}R(\theta)
-
\inf_{\theta\in\M_k}R_{E,n}(\theta)
>
\sqrt{k}\delta+\vep
\right)
\\
&\leqslant
\underbrace{
\pp\left(
\sup_{\theta\in\M_0}
(R(\theta)-R_{E,n}(\theta))>\vep
\right)
}_{(\Pi_0)}
\\
&\quad+
\sum_{k=1}^{M}
\underbrace{
\pp\left(
\sup_{\theta\in\M_k}
(R(\theta)-R_{E,n}(\theta))
>
\sqrt{k}\delta+\vep
\right)
}_{(\Pi_k)}.
\endaligned
\nneq
For each error probability $(\Pi_k)$, we will apply Bousquet's inequality.

To this end, we first estimate the bias on each risk-level subclass $\M_k$. For every $0\leqslant k\leqslant M$ such that $\M_k\neq\emptyset$, set
$$
D_k:={\rm Diam}(\M_k,d_{\psi_1}).
$$
Then Lemma \ref{lem31} gives
 \bbeq\label{320}
 \aligned
 \ee &\sup_{\theta\in\M_k} (R (\theta) - R_{E,n}(\theta)) \\
 &\leqslant 12 \sqrt{\frac{2}{n}} \int_{0}^{D_k/2} \sqrt{\log \mathcal{P}(\M_k, d_{\psi_1},\vep)} d\vep + \frac{12}{n} \int_{0}^{D_k/2} \log \mathcal{P}(\M_k, d_{\psi_1},\vep) d\vep\\
 &=:b(n,\M_k).
 \endaligned
 \nneq
If $\M_k=\emptyset$, we set
$$
D_k:=0,\qquad b(n,\M_k):=0.
$$

Set
\bbeq\label{p+estimateLb}
\eta_k(n,x)
:=
\sqrt{
\frac{
2\left(
\sigma^2(\Theta)
+
2(L+R^*)b(n,\M_k)
\right)x
}{n}
}
+
\frac{(L+R^*)x}{3n}
\nneq
which is the random fluctuation bound given in Bousquet's inequality.  

\bthm\label{thmM26}{\bf (upper bound for the minimal risk)}
Assume {\bf (H1)} and {\bf (H2)}. Suppose moreover the lower boundedness condition
\eqref{thmM26a}, and assume that $b(n,\M_k)$ are finite for all $0\leqslant k\leqslant M$.

Then, for every $n\in \nn^*, \delta>0$ and $x>0$, provided that
\bbeq\label{thmM26bb}
\sqrt{k}\delta
\geqslant
b(n,\M_k)-b(n,\M_0)+\eta_k(n,kx)-\eta_0(n,x),
\qquad 1\leqslant k\leqslant M,
\nneq
the following non-asymptotic estimate holds: for
$$
\vep:=\vep(n,\delta,x)=b(n,\M_0)+\eta_0(n,x),
$$
we have
\bbeq\label{p+estimateL}
p_+(n,\vep)
=
\pp\left(
\inf_{\theta\in\Theta}R(\theta)
>
\inf_{\theta\in\Theta}R_{E,n}(\theta)+b(n,\M_0)+\eta_0(n,x)
\right)
\leqslant
e^{-x}+\frac{e^{-x}}{1-e^{-x}}.
\nneq
\nthm

\brmk\label{38} {\rm
Let
$$
d_B:=\dim_B(\Theta,d_{\psi_1}).
$$
Then, for some constant $C>0$,
$$
\log\left(1+\mathcal P(\Theta,d_{\psi_1},s)\right)
\leqslant
C(1+d_B)\log\frac1s,
\qquad s\in(0,1/e).
$$
Assume that 
\bbeq\label{rmk38a}
n \gtrsim \frac{1+d_B}{\delta^2},
\nneq
and fix $x>0$ such that $e^{-x}+\frac{e^{-x}}{1-e^{-x}} \leqslant \alpha\in (0,1/2)$, where $1-\alpha$ is the fixed confidence level. As $\max_{k\leqslant M} b(n,\M_k)\lesssim \sqrt{\frac{1+d_B}{n}}$ and $\eta_k(n, kx)\lesssim \sqrt{\frac{kx}{n}}$, the risk-level condition \eqref{thmM26bb} is satisfied for all $k\geqslant N_0$ where $N_0$ is independent of $\delta$. 

For $k=0,\ldots,N_0$, suppose that the minimizer set $\M$ of the true
risk function $R$ is a singleton and that, in the typical case,
$D_k \lesssim \delta^{\beta}$ for $k=0,\cdots, N_0$ and for some $\beta>0$. Then, for all sufficiently small $\delta>0$,
$$
b(n,\M_k)
\lesssim
\sqrt{\frac{1+d_B}{n}}
\delta^{\beta}\log\frac{1}{\delta}
\lesssim
\delta^{1+\beta}\log\frac{1}{\delta}
$$
by Lemma \ref{lem31} and condition \eqref{rmk38a}. The
condition \eqref{thmM26bb} is satisfied for all $k\geqslant1$ and for all $\delta\in(0,\delta_0]$, where $\delta_0\leqslant 1/e$ is
sufficiently small. The same conclusion remains valid when
$\M$ is finite, with a little more effort.

Consequently under the condition \eqref{rmk38a}, Theorem \ref{thmM26} gives
$$
p_+(n,\vep(n,\delta,x))
\leqslant
e^{-x}+\frac{e^{-x}}{1-e^{-x}}\leqslant \alpha
$$
where 
\bbeq\label{rmk38b}
\vep(n,\delta,x)
\lesssim \sqrt{\frac{1+d_B}{n}}  \delta^\beta \log \frac{1}{\delta} + 
\sqrt{
\frac{
x\left[
\sigma^2(\Theta)+C(1+d_B)\delta^{1+\beta}\log(1/\delta)
\right]
}{n}
}.
\nneq

}
\nrmk

\brmk
{\rm
Let us compare Theorem \ref{thmM26} with the localized excess-risk bounds in Koltchinskii
\cite[Chapters 1 and 4]{Kol11}. The  results therein concern the excess risk of
an empirical minimizer,
$$
R(\hat\theta_n)-R_*,
\qquad
\hat\theta_n\in\argmin_{\theta\in\Theta}R_{E,n}(\theta),
$$
and localize the empirical process on risk-level sets. For
$r>0$, put
$$
\Theta_r:=\{\theta\in\Theta:R(\theta)-R_*\leqslant r\},
\qquad
\mathcal Q(r):=\{Q_\theta-Q_{\theta'}:\theta,\theta'\in\Theta_r\}.
$$
The localized complexity and diameter are then of the form
$$
\phi_n(r)
:=
\ee\sup_{h\in\mathcal Q(r)}
\left|(L_n-\mu)(h)\right|,
\qquad
D_{\rho_\mu}(r)
:=
\sup_{\theta,\theta'\in\Theta_r}
\rho_\mu(Q_\theta,Q_{\theta'}),
$$
where $\rho_\mu$ is a pseudo-metric on $L^2(\mu)$ such that
$$
\rho_\mu^2(f,g)
\geqslant
{\rm Var}_\mu(f-g),
\qquad f,g\in L^2(\mu).
$$ 
The resulting scale is
described by a fixed-point argument for finding the maximal risk level $r^*$ satisfying the inequality
$$
r
\leqslant 
\phi_n(r)
+
\sqrt{
\frac{x}{n}
\left(
D_{\rho_\mu}^2(r)+2\phi_n(r)
\right)}
+
\frac{x}{n}.
$$
The results in \cite[Chapters 1 and 4]{Kol11} on the excess-risk problem give the non-asymptotic estimate of $\pp(\hat{\theta}_n \in \Theta_{r^*})=\pp(R(\hat{\theta}_n)-R_*\leqslant r^*)$. 

By contrast, Theorem \ref{thmM26} estimates the deviation probability
$p_+(n,\vep)$ of the minimal empirical risk, which is a different problem. Our prerequisite condition  \eqref{thmM26bb} is inspired by the excess-risk results in  \cite[Chapter 4]{Kol11}. 
The present approach gives a direct non-asymptotic estimate for the deviation of the minimal
empirical risk, without solving a fixed-point equation.

For related progress on algorithmic stability, see Bousquet--Elisseeff \cite{BE02}
and Bousquet {\it et al.} \cite{BousquetKZ20}. For concentration inequalities for
empirical minimizers, see Escande \cite{Esc23} and the references therein.
}
\nrmk

\subsection{Talagrand's concentration inequality for the upper bound of the minimal risk in the exponentially integrable case}

In this paragraph we remove the lower boundedness assumption of the risk functions $Q_\theta$ in Theorem \ref{thmM26}.  We impose the following envelope condition on the negative part of 
the risk function $Q_\theta$:

\medskip
{\bf (H3)} {\it  There exists a measurable function $Q^*:S\to[0,+\infty)$ such that
$$
Q_\theta(z)^{-}=\max\{-Q_\theta(z),0\}\leqslant Q^*(z),
\qquad \forall z\in S,\ \forall \theta\in\Theta,
$$
($Q^*$ is an envelope of $\{Q_\theta^-; \ \theta\in \Theta\}$) and
$$
\|Q^*\|_{\psi_1}<+\infty.
$$

}
\medskip

For the unbounded case, we use the same risk-level sets $\M_k$, $0\leqslant k\leqslant M$,
and the entropy bias bounds $b(n,\M_k)$ defined in \eqref{320}.

For $L>0$, define
$$
Q_\theta^L:=(-L)\vee Q_\theta,
\quad
R^L(\theta):=\mu(Q_\theta^L),
\quad
R_*^L:=\inf_{\theta\in\Theta}R^L(\theta),
\quad
\chi_L:=Q^*\mathbf 1_{\{Q^*>L\}}.
$$
For $x>0$, set
\bbeq\label{tau-tail-312}
\tau_n(x,L)
:=
\mu(\chi_L)
+
2\|\chi_L\|_{\psi_1}\sqrt{\frac{x}{n}}
+
\|\chi_L\|_{\psi_1}\frac{x}{n}.
\nneq

For fixed $\delta\in(0,1]$, $0\leqslant k\leqslant M$ and $x>0$, set
\bbeq\label{eta1-L}
\eta_{k,L}(n,x)
:=
\sqrt{
\frac{
2\left(
\sigma^2(\Theta)
+
2\left(R^*+L+\frac{\delta}{4}\right)b(n,\M_k)
\right)x
}{n}
}
+
\frac{
\left(R^*+L+\frac{\delta}{4}\right)x
}{3n}.
\nneq

\bthm\label{thmM39}
Assume {\bf (H1)}, {\bf (H2)} and {\bf (H3)}. Let
$\delta\in(0,1]$, $L>0$ and $x>0$, and assume that
$b(n,\M_k)$ is finite for every $0\leqslant k\leqslant M$.
Suppose that
\bbeq\label{thmM39a}
2(L+\| Q^*\|_{\psi_1})\exp\left(-\frac{L}{\| Q^*\|_{\psi_1}}\right) \leqslant \frac{\delta}{4}
\nneq
(which implies
$
\mu(\chi_L)
\leqslant
\frac{\delta}{4}
$)
and that the peeling condition
\bbeq\label{thmM39-layer}
\sqrt{k}\delta-\frac{\delta}{4}
\geqslant
b(n,\M_k)-b(n,\M_0)
+
\eta_{k,L}(n,kx)-\eta_{0,L}(n,x),
\qquad
1\leqslant k\leqslant M,
\nneq
holds.
Then, for
\bbeq\label{vep12}
\vep
=
\vep(n,\delta,x,L)
:=
2\max\left\{
b(n,\M_0)+\eta_{0,L}(n,x),
\,
\tau_n(x,L)
\right\},
\nneq
we have
\bbeq\label{vep11}
p_+(n,\vep)
=
\pp\left(
\inf_{\theta\in\Theta}R(\theta)>
\inf_{\theta\in\Theta}R_{E,n}(\theta)+\vep
\right)
\leqslant
2e^{-x}
+
\frac{e^{-x}}{1-e^{-x}}.
\nneq
\nthm

\brmk\label{adamczak}
{\rm
Let us compare Theorem \ref{thmM39} with Adamczak's  Talagrand
inequality \cite[Theorem 4]{Ada08} in the unbounded case. For $\alpha\in (0,1]$, $\psi_\alpha=e^{|x|^\alpha}-1$ (which is not convex when $0<\alpha<1$), define $\|X\|_{\psi_\alpha}$ in the same way. If 
$$
\left\|\sup_{\theta\in \Theta} |Q_\theta(Z)|\right\|_{\psi_\alpha}<+\infty, 
$$ 
letting $W_n:= \sqrt{n}\sup_{\theta\in \Theta} |R_{E,n}(\theta) -R(\theta)|$, then for any $\eta, \delta\in (0,1)$, there is some constant $C=C(\alpha, \eta,\delta)>0$ such that for 
any $n\geqslant 1, r>0$
$$
\aligned
\pp\left( W_n > (1+\eta)\ee W_n + r \right) \leqslant &\exp\left( - \frac{r^2}{2(1+\delta) \sigma^2(\Theta)}\right) \\
&+ 3 \exp\left( - \left(\frac{r\sqrt{n}}{ C \left\|\max_{i\leqslant n} \sup_{\theta\in \Theta} |Q_\theta(Z_i)|\right\|_{\psi_\alpha}}\right)^{\alpha}\right). 
\endaligned
$$

 Adamczak's result has the advantage of being
applicable under the more general $\psi_\alpha$-integrability condition, with
$\alpha\in(0,1]$, and hence includes tails heavier than the exponential case when
$\alpha<1$.

In the present $\psi_1$ framework, however, the truncation argument gives a
slightly sharper tail treatment when $\alpha=1$. Indeed, after setting
$$
Q_\theta^L:=(-L)\vee Q_\theta,
\qquad
\chi_L:=Q^*\mathbf 1_{\{Q^*>L\}},
$$
we have
$$
0
\leqslant
Q_\theta^L-Q_\theta
\leqslant
\chi_L.
$$
Thus the tail part is controlled by a scalar Bernstein inequality for the single
observable $\chi_L$, rather than by an unbounded empirical-process inequality. This
gives the remainder
$$
\tau_n(x,L)
=
\mu(\chi_L)
+
2\|\chi_L\|_{\psi_1}\sqrt{\frac{x}{n}}
+
\|\chi_L\|_{\psi_1}\frac{x}{n}.
$$

In particular, the remainder tail term is only
$$
\|\chi_L\|_{\psi_1}\frac{x}{n}.
$$

By contrast, applying Adamczak's inequality directly to the centered tail class
would involve a term of the form
$$
\frac{x}{n}
\left\|
\max_{1\leqslant i\leqslant n}
|\chi_L(Z_i)-\mu(\chi_L)|
\right\|_{\psi_1},
$$
which is bounded, up to a universal constant, by
$$
\|\chi_L\|_{\psi_1}\frac{x\log(1+n)}{n}.
$$
Hence, in the exponential-integrability case $\alpha=1$, under {\bf (H3)}, our
truncation argument controls the tail by the single variable $\chi_L$ and avoids
the extra factor $\log(1+n)$, while Adamczak's inequality remains more general for
$\psi_\alpha$ tails with $\alpha\in(0,1]$.
}
\nrmk

\section{Proof of Theorem \ref{thm_Ma}}

We now prove Theorem \ref{thm_Ma}, which is crucial for our estimates of $1-p_-(n,\vep)$ on the lower bounds of the true risk, in Theorems \ref{thmM21} and \ref{thmM22}.

\bprf[Proof of Theorem \ref{thm_Ma}]
\textbf{(1).} By Theorem \ref{thm_GL}, it is enough to prove that
  \bbeq \label{3thmBern2d-1}
  \int_S fd(\nu-\mu) \leqslant \sqrt{2 H(\nu|\mu)\left( \var_\mu(f) + \frac{1}{3} L(\vep)\right) } + \sqrt{\frac{2}{3\vep}}  H(\nu|\mu), \ \forall \nu.
  \nneq
  Below we follow the elementary and beautiful method of \cite[Proof of Theorem 1(ii)]{BolleyVillani05}, with some necessary changes for our purpose.

 Without loss of generality assume that $d\nu=h d\mu=(1+u) d\mu$ ($u=h-1$) and $\mu(f)=0$. Setting
 $$
 \varphi(u)= (1+u) \log (1+u) - u,
 $$
 we have by the second-order Taylor's formula with integral remainder
$$
  H(\nu|\mu) = \int_S \varphi(u) d\mu = \int_S \int_{0}^{1} \frac{u^2(x) (1-t)}{1+t u(x)} dt \mu(dx).
  $$
We have by Cauchy-Schwarz inequality
$$
\aligned
&\int_{0}^{1}(1-t) dt \int_S fd(\nu-\mu) =\int_S\int_{0}^{1} (1-t) f u dt d\mu\\
& \leqslant \sqrt{\int_S\int_{0}^{1} \frac{u^2(x) (1-t)}{1+t u(x)} dt \mu(dx) \int_S\int_{0}^{1} (1-t)(1+t u(x)) f^2(x) dt \mu(dx)  }\\
&=\sqrt{H C}
\endaligned
$$
where $H=H(\nu|\mu)$ and
$$
\aligned
C&= \int_S\int_{0}^{1} (1-t)(1+t u(x))  f^2(x)dt \mu(dx)\\
&= \frac{1}{2} \mu(f^2) + \frac{1}{6} \int_S u(x) f^2 \mu(dx)\\
&=\frac{1}{2} \mu(f^2) + \frac{1}{6} \int_S  (f^2-\mu(f^2)) h \mu(dx).\\
\endaligned
$$
By the variational formula of relative entropy,
$$
\aligned
\int_S (f^2-\mu(f^2)) h \mu(dx) &\leqslant \frac{1}{\vep}\left( H(\nu|\mu) + \log \int_S e^{\vep(f^2-\mu(f^2))} d\mu\right)\\
&=L(\vep) + \frac{1}{\vep} H.
\endaligned
$$
Substituting those two estimates into the previous inequality, we obtain
\bbeq\label{2.2}
\aligned
\int_S fd(\nu-\mu) &\leqslant \sqrt{4H\left(\frac{1}{2} \var_\mu(f) + \frac{1}{6} L(\vep) + \frac{1}{6\vep} H\right) }\\
&\leqslant \sqrt{2 H\left( \var_\mu(f) + \frac{1}{3} L(\vep)\right) } + \sqrt{\frac{2}{3\vep}} H
\endaligned
\nneq
which is the desired \eqref{3thmBern2d-1} (where we have applied the elementary inequality $\sqrt{a+b}\leqslant \sqrt{a}+ \sqrt{b}$ for $a,b\geqslant 0$).

The bound \eqref{thm_M3} of $L(\vep)$ follows from Theorem \ref{thm_BV}.

Combining \eqref{thm_M3} and the first inequality in \eqref{2.2}, and optimizing over $\vep\leqslant \frac{1}{2\|\tilde{f}^2-\mu(\tilde{f}^2)\|_{\psi_1}}$, we get: if  $H=H(\nu|\mu)\leqslant \frac{1}{2}$,
\bbeq\label{TI-f}
\aligned
\int_S fd(\nu-\mu) &\leqslant \sqrt{4H\left(\frac{1}{2} \var_\mu(f) + \frac{\sqrt{2}\|\tilde{f}^2-\mu(\tilde{f}^2)\|_{\psi_1} }{3}\sqrt{H}\right)}\\
&\leqslant \sqrt{2 \var_\mu(f) H} + 2\sqrt{\frac{\sqrt{2}\|\tilde{f}^2-\mu(\tilde{f}^2)\|_{\psi_1} }{3}} H^{3/4}.
\endaligned
\nneq
This transport-entropy inequality implies \eqref{thm_M3b}, by Gozlan-L\'eonard \cite[Theorem 2]{GL07}.

\textbf{(2).}
Approximating $f$ by $\max\{f\wedge L, -L\}$ with $L\to+\infty$, we may assume without loss of
generality that $f$ is exponentially integrable.

We will combine the well-known Theorem \ref{thmA} with Theorem \ref{thm_BV}. For any $L>0$, let $f_L=f\wedge L=\min\{f,L\}$. Notice that $\var_\mu(f_L)\leqslant \var_\mu(f)$.
By Theorem \ref{thmA} and Theorem \ref{thm_GL},
for any $\nu\in M_1(S)$ such that $H=H(\nu|\mu)<+\infty$ and $\nu(|f|)<+\infty$,
$$
\nu(f_L)-\mu(f_L) \leqslant \sqrt{2\var_\mu(f_L) H} + \frac{L}{3}H \leqslant \sqrt{2\var_\mu(f) H} + \frac{L}{3}H.
$$
On the other hand by Theorem \ref{thm_BV},
$$
\nu(f-f_L)-\mu(f-f_L) \leqslant \|f-f_L\|_{\psi_1} \left(2 \sqrt{H} + H\right).
$$
Writing $\vep(L)= \|f-f_L\|_{\psi_1}$ and taking the sum of the above two inequalities, we obtain
$$
\aligned
&\nu(f)- \mu(f) \leqslant \left(\sqrt{2\var_\mu(f)} + 2 \vep(L)\right) \sqrt{H} +  \left(\frac{L}{3}+ \vep(L)\right) H\\
&\leqslant \sqrt{2 \left(\var_\mu(f)+ 2\sqrt{2 \var_\mu(f)} \vep(L)+ 2 \vep(L)^2 \right) H} + \left(\frac{L}{3}+ \vep(L)\right) H
\endaligned
$$
which implies the desired \eqref{thm_M4}, by Theorem \ref{thm_GL}.
\nprf

\section{Proofs of the upper bounds}

\subsection{Proof of Theorem \ref{thmM23}}

We begin with a classic result.

\bthm\label{lem41} (\cite[Lemma 2.2.10]{vdVW96}) Let $X_i, 1\leqslant i\leqslant m$ be real random variables satisfying the Bernstein's inequality:
$$
\pp(|X_i|> x) \leqslant 2 \exp\left(-\frac{x^2}{2(c_B + M x)}\right),
$$
for all $x$ (and $i$) and fixed $c_B, M > 0$. Then there are universal constants $K_1, K_2>0$ such that
\bbeq\label{5Eq1}
\|\max_{1\leqslant i\leqslant m} |X_i|\|_{\psi_1} \leqslant K_1 \sqrt{c_B \log (1+m)} +  K_2 M \log(1+m).
\nneq
\nthm

In \cite[Lemma 2.2.10]{vdVW96}, $K_1, K_2$ are two absolute constants, not explicit. We will prove  \eqref{5Eq1} with the explicit constants
\bbeq\label{thmM23ba}
K_1=\frac{2\sqrt{3}}{\log 2},\qquad K_2= \frac{12}{\log 2}.
\nneq
in the Appendix.

We are ready to prove Theorem \ref{thmM23}, by means of the important "{\it chaining}" technique combined with the Bernstein's inequality in Theorem \ref{thm_BV}.

\begin{proof}[Proof of Theorem \ref{thmM23}]

First, without loss of generality we may and will assume that $\Theta$ is finite. We divide its proof into 5 steps.

{\bf Step 1.}
By the i.i.d. property of $(Z_k)$,  we have
\bbeq\label{51I}
\log \ee e^{\lambda W_n(\theta_0)} \leqslant \frac{\lambda^2 c_B}{2\left(1 - \frac{M}{\sqrt{n}} |\lambda|\right)}, \ \text{ if } |\lambda|< \frac{\sqrt{n}}{M}
\nneq
where $c_B=c_B(\theta_0), M=M(\theta_0)$ are the constants in the Bernstein's inequality satisfied by $\pm \tilde{Q}_{\theta_0}$ (both). We may assume that $\tilde{Q}_{\theta_0}$ is not zero a.s..

If $\|W_n(\theta_0)\|_{\psi_1} \geqslant  \frac{4M}{\sqrt{n}}$, $\lambda_0:=\|W_n(\theta_0)\|_{\psi_1}^{-1}$ verifies $2\lambda_0 \leqslant \frac{\sqrt{n}}{2M}$ and then
$$
\aligned
2\log 2 &\leqslant \log \ee e^{2\lambda_0 |W_n(\theta_0)|} \text{ (Cauchy-Schwarz)}\\
&\leqslant \log \left(\ee e^{2\lambda_0 W_n(\theta_0)} + \ee e^{-2\lambda_0 W_n(\theta_0)}\right)\\
&\leqslant \log 2 +  4\lambda_0^2 c_B \text{\ \  ( by \eqref{51I} )}
\endaligned
$$
which yields: once $\|W_n(\theta_0)\|_{\psi_1} \geqslant  \frac{4M}{\sqrt{n}}$,
\bbeq
\|W_n(\theta_0)\|_{\psi_1} \leqslant 2 \sqrt{\frac{c_B}{\log 2}} .
\nneq
That proves \eqref{thmM23i} in part (1) of Theorem \ref{thmM23}.

Applying \eqref{thmM23i} with $c_B=2 \|Q_{\theta_0}\|_{\psi_1}^2, M=\|Q_{\theta_0}\|_{\psi_1}$ (by Theorem \ref{thm_BV}), we obtain \eqref{thmM23i2}, because for $n\geqslant 2> 2\log 2$, the first term in the maximum of  \eqref{thmM23i} is larger than the second one. The proof of part (1) is completed.

\medskip
{\bf Step 2.} By Bolley-Villani and Gozlan-L\'eonard's Theorem \ref{thm_BV},
\bbeq\label{eq52}
\pp\left(W_n(\theta_1)- W_n(\theta_2)> d_{\psi_1}(\theta_1,\theta_2) \left(2\sqrt{x} + \frac{x}{\sqrt{n}} \right)   \right) \leqslant e^{-x},\ x>0
\nneq
which implies
$$
\pp\left(|W_n(\theta_1)- W_n(\theta_2)|> x\right) \leqslant 2 \exp\left(-\frac{x^2}{2(c_B + M x)}\right), x>0
$$
with
$$
c_B=2 d_{\psi_1}(\theta_1,\theta_2)^2, \qquad M= \frac{d_{\psi_1}(\theta_1,\theta_2)}{\sqrt{n}}.$$
By Theorem \ref{lem41}, we obtain: for any $\vep>0$ and for any $m$ couples $(\theta_1(i), \theta_2(i)), 1\leqslant i\leqslant m$ such that $d_{\psi_1}(\theta_1(i), \theta_2(i))\leqslant \vep$,
\bbeq\label{eq53}
\|\max_{1\leqslant i\leqslant m} |W_n(\theta_1(i))- W_n(\theta_2(i)) |\|_{\psi_1} \leqslant  K_1 \vep \sqrt{ 2 \log (1+m)} + \frac{K_2}{\sqrt{n}} \vep \log(1+m).
\nneq

\medskip

{\bf Step 3 (chaining, proof of \eqref{thmM23a}).}   For any $\eta>0$, let us set the dyadic scale
$$
\vep_k = \eta 2^{-k}, \ k\in\nn
$$
and nested subsets $\Theta_0\subset \Theta_1\subset\cdots \subset \Theta_{K}=\Theta$ such that each $\Theta_k$ is  $\vep_k$-separated (in the metric $d_{\psi_1}$) and maximal in the sense that if one adds one point $\theta$ to $\Theta_k$,
$\Theta_k\bigcup \{\theta\}$ is no longer $\vep_k$-separated. By the definition of packing number, the number $\#(\Theta_k)$  of points in $\Theta_k$ verifies
$$
\#(\Theta_k)\leqslant {\mathcal P}(\Theta, d_{\psi_1}, \vep_k).
$$
For every $\theta_{k+1}\in \Theta_{k+1}$, link it to a single point  $\theta_{k}\in \Theta_k$ such that $d_{\psi_1}(\theta_{k+1}, \theta_k)\leqslant \vep_k$. Then each $\theta_{k+1}\in \Theta_{k+1}$ is connected to a point $\theta_0\in \Theta_0$ by a chain $\theta_{k+1}, \theta_k,\cdots, \theta_0$. Thus  for each $\theta=\theta_K\in \Theta_K=\Theta$, we have the following telescopic sum
\bbeq\label{53b}
W_n(\theta) = W_n(\theta_0) + \sum_{k=0}^{K-1} \left(W_n(\theta_{k+1}) - W_n(\theta_k)\right).
\nneq
Thus we obtain 
\bbeq\label{53c}
\sup_{\theta\in \Theta} |W_n(\theta)| \leqslant \max_{\theta_0\in \Theta_0} |W_n(\theta_0)| + \sum_{k=0}^{K-1} \sup_{\theta_{k+1}\in \Theta_{k+1}} |W_n(\theta_{k+1}) - W_n(\theta_k)|.
\nneq
Since  $d_{\psi_1}(\theta_{k+1}, \theta_k)\leqslant \vep_k$, by the key inequality \eqref{eq53} in Step 2, we get
\bbeq\label{eq54}
\aligned
&\|\sup_{\theta_{k+1}\in \Theta_{k+1}} |W_n(\theta_{k+1}) - W_n(\theta_k)|\|_{\psi_1} \\
&\leqslant  K_1 \vep_k \sqrt{ 2 \log (1+\#(\Theta_{k+1}))} + \frac{K_2}{\sqrt{n}} \vep_k \log(1+\#(\Theta_{k+1}))\\
& \leqslant K_1 \vep_k \sqrt{ 2 \log (1+{\mathcal P}(\Theta, d_{\psi_1}, \vep_{k+1}))} + \frac{K_2}{\sqrt{n}} \vep_k \log(1+{\mathcal P}(\Theta, d_{\psi_1}, \vep_{k+1})).
\endaligned
\nneq
Since for a non-increasing function $f$ on $(0,+\infty)$,
$$
\int_{\vep_{k+2}}^{\vep_{k+1}} f(\vep) d\vep \geqslant f(\vep_{k+1}) (\vep_{k+1}-\vep_{k+2})= \frac{1}{4} f(\vep_{k+1}) \vep_k
$$
we obtain from \eqref{eq54}

\bbeq\label{eq55}
\aligned
&\sum_{k=0}^{K-1}\|\sup_{\theta_{k+1}\in \Theta_{k+1}} |W_n(\theta_{k+1}) - W_n(\theta_k)|\|_{\psi_1}\\
 &\leqslant 4 \sqrt{ 2} K_1 \int_{0}^{\eta} \sqrt{ \log (1+{\mathcal P}(\Theta, d_{\psi_1}, \vep))} d\vep + \frac{4K_2}{\sqrt{n}} \int_{0}^{\eta} \log(1+{\mathcal P}(\Theta, d_{\psi_1}, \vep)) d\vep.
\endaligned
\nneq
Now taking $\eta=D={\rm Diam}(\Theta, d_{\psi_1})$, and beginning the nested subsets with $\Theta_0$ being a singleton $\{\theta_0\}$ ($\theta_0$ is an arbitrary fixed point), we obtain \eqref{thmM23a} by
 the inequality above and  \eqref{53c}. We have so proved part (2) of Theorem \ref{thmM23}.

\medskip

{\bf Step 4 (proof of \eqref{thmM23b2}).} In Step 3, taking $\eta=D={\rm Diam}(\Theta, d_{\psi_1})$, $\Theta_0$ is reduced to a singleton $\{\theta_0\}$. Instead of \eqref{53c}, we have by \eqref{53b},
$$
\sup_{\theta\in \Theta} |W_n(\theta)| \leqslant |W_n(\theta_0)| + \sum_{k=0}^{K-1} \sup_{\theta_{k+1}\in \Theta_{k+1}} |W_n(\theta_{k+1}) - W_n(\theta_k)|
$$

Since $\ee W_n(\theta_0)=0$ and $\ee |W_n(\theta_0)|  \leqslant \sqrt{\var(W_n(\theta_0))}= \sigma(\theta_0)$, we obtain \eqref{thmM23b2} by \eqref{eq55} (as in Step 3). We have so completed the proof of part (3) of Theorem \ref{thmM23}.
\medskip

{\bf Step 5 (proof of \eqref{thmM23c} in part (4): forward and backward double chaining  by following \cite[Proof of Theorem 2.2.6]{vdVW96}).}  In the chaining in Step 3, for any $\eta>0$ and for any
$\theta=\theta_K, \theta'=\theta'_K\in \Theta_K=\Theta$ , let
$$
\theta_K, \cdots, \theta_0;\ \theta'_K, \cdots, \theta'_0;
$$
be the two chains linking $\theta_K, \theta'_K$ to $\theta_0,  \theta'_0$ in $\Theta_0$. Since
$$
\aligned
&|(W_n(\theta)- W_n(\theta_0))- (W_n(\theta')- W_n(\theta'_0))|\\
&\leqslant \sum_{k=0}^{K-1} |(W_n(\theta_{k+1})- W_n(\theta_k))| + \sum_{k=0}^{K-1} |(W_n(\theta'_{k+1})- W_n(\theta'_k))|,
\endaligned
$$
by \eqref{eq55}, we obtain
\bbeq\label{58}
\aligned
\|&\max_{\theta, \theta'\in \Theta} |(W_n(\theta)- W_n(\theta_0))- (W_n(\theta')- W_n(\theta'_0))|\|_{\psi_1}\\
&\leqslant 2  \sum_{k=0}^{K-1}\|\sup_{\theta_{k+1}\in \Theta_{k+1}} |W_n(\theta_{k+1}) - W_n(\theta_k)|\|_{\psi_1}\\
&\leqslant 8 \sqrt{ 2} K_1 \int_{0}^{\eta} \sqrt{ \log (1+{\mathcal P}(\Theta, d_{\psi_1}, \vep))} d\vep + \frac{8K_2}{\sqrt{n}} \int_{0}^{\eta} \log(1+{\mathcal P}(\Theta, d_{\psi_1}, \vep)) d\vep.
\endaligned
\nneq
Then
$$
\|\max_{\theta, \theta'\in \Theta: d_{\psi_1}(\theta,\theta')\leqslant \delta} |W_n(\theta)- W_n(\theta')|\|_{\psi_1}
$$
is bounded by the r.h.s. of \eqref{58} plus
$
\|\max_{\theta_0, \theta'_0} |W_n(\theta_0)- W_n(\theta'_0)|\|_{\psi_1},
$
where the maximum is taken over all pairs of endpoints of the chains starting from $\theta, \theta'$ with $d_{\psi_1}(\theta,\theta')\leqslant \delta$. For every such pair of endpoints
$\theta_0, \theta'_0$, choose exactly one $(\theta_K, \theta'_K)$ with $d_{\psi_1}(\theta_K,\theta'_K)\leqslant \delta$ such that the chains starting from $\theta_K, \theta'_K$ end at
$\theta_0, \theta'_0$ respectively. The cardinality of the set $\Theta(\delta)$ of all such pairs $(\theta_K, \theta'_K)$ is $\leqslant \#(\Theta_0)^2\leqslant {\mathcal P}(\Theta, d_{\psi_1}, \eta)^2$. Thus by \eqref{eq53} in Step 2,
$$
\aligned
&\|\max_{(\theta_K, \theta'_K)\in \Theta(\delta)} |W_n(\theta_K)- W_n(\theta'_K)|\|_{\psi_1} \\
&\leqslant  K_1 \delta \sqrt{ 2 \log (1+{\mathcal P}(\Theta, d_{\psi_1}, \eta)^2)} + \frac{K_2}{\sqrt{n}} \delta \log(1+{\mathcal P}(\Theta, d_{\psi_1}, \eta)^2).
\endaligned
$$
Using this control and the elementary inequality
$$
\aligned
|W_n(\theta_0)- W_n(\theta'_0)| \leqslant |(W_n(\theta_K)- W_n(\theta_0))- (W_n(\theta'_K)- W_n(\theta'_0))| + |W_n(\theta_K)- W_n(\theta'_K)|
\endaligned
$$
we obtain by \eqref{58},
\bbeq\label{59}
\aligned
&\|\max_{\theta, \theta'\in \Theta: d_{\psi_1}(\theta,\theta')\leqslant \delta} |W_n(\theta)- W_n(\theta')|\|_{\psi_1}\\
&\leqslant 2\left(8 \sqrt{ 2} K_1 \int_{0}^{\eta} \sqrt{ \log (1+{\mathcal P}(\Theta, d_{\psi_1}, \vep))} d\vep + \frac{8K_2}{\sqrt{n}} \int_{0}^{\eta} \log(1+{\mathcal P}(\Theta, d_{\psi_1}, \vep)) d\vep\right)\\
&\quad +  K_1 \delta \sqrt{ 2 \log (1+{\mathcal P}(\Theta, d_{\psi_1}, \eta)^2)} + \frac{K_2}{\sqrt{n}} \delta \log(1+{\mathcal P}(\Theta, d_{\psi_1}, \eta)^2)
\endaligned
\nneq
which gives us the desired \eqref{thmM23c}.

The proof of Theorem \ref{thmM23} is completed.
\end{proof}

\subsection{Proof of Theorem \ref{thmM26}}

We begin by estimating the bias terms.

\blem\label{lem51}
Assume {\bf (H1)} and {\bf (H2)}. For $0\leqslant k\leqslant M$ such that $\M_k$ is non-empty, let
$$
D_k:={\rm Diam}(\M_k,d_{\psi_1}).
$$
Then
\bbeq\label{lem51a}
\aligned
\ee\sup_{\theta\in\M_k}
\left(R(\theta)-R_{E,n}(\theta)\right)
&\leqslant
12\sqrt{\frac{2}{n}}
\int_0^{D_k/2}
\sqrt{\log\mathcal P(\M_k,d_{\psi_1},\vep)}\,d\vep
\\
&\quad+
\frac{12}{n}
\int_0^{D_k/2}
\log\mathcal P(\M_k,d_{\psi_1},\vep)\,d\vep.
\endaligned
\nneq
\nlem

\bprf[Proof of Lemma \ref{lem51}]

Assume that $\M_k\neq\emptyset$.
Applying Lemma \ref{lem31} to the class
$$
\{-Q_\theta:\theta\in\M_k\},
$$
we obtain
\bbeq
\aligned
\ee\sup_{\theta\in\M_k}
\left(R(\theta)-R_{E,n}(\theta)\right)
&=
\ee\sup_{\theta\in\M_k}
\left(L_n(-Q_\theta)-\mu(-Q_\theta)\right)
\\
&\leqslant
12\sqrt{\frac{2}{n}}
\int_0^{D_k/2}
\sqrt{\log\mathcal P(\M_k,d_{\psi_1},\vep)}\,d\vep
\\
&\quad+
\frac{12}{n}
\int_0^{D_k/2}
\log\mathcal P(\M_k,d_{\psi_1},\vep)\,d\vep,
\endaligned
\nneq
where
$$
D_k={\rm Diam}(\M_k,d_{\psi_1}).
$$
This proves \eqref{lem51a}.
\nprf

\bprf[Proof of Theorem \ref{thmM26}]
We estimate the probabilities $(\Pi_0)$ and $(\Pi_k)$, $1\leqslant k\leqslant M$,
in the decomposition \eqref{thmM26b}.

By the lower boundedness assumption \eqref{thmM26a}, for every $\theta\in\Theta$,
$$
-Q_\theta\leqslant L.
$$
Moreover, since $R(\theta)\leqslant R^*$ for every $\theta\in\Theta$, we have
$$
R(\theta)-Q_\theta\leqslant L+R^*.
$$
Thus, on each class $\M_k$, the one-sided boundedness constant $b$ in Bousquet's
inequality can be chosen as $L+R^*$.

For every $0\leqslant k\leqslant M$ such that $\M_k$ is non-empty, applying Bousquet's inequality in Theorem
\ref{Tala_concentration} to the class
$$
\left\{
R(\theta)-Q_\theta:\theta\in\M_k
\right\},
$$
and using Lemma \ref{lem51}, we obtain, for every $y>0$,
\bbeq\label{layer-bound}
\aligned
&\pp\left(
\sup_{\theta\in\M_k}
\left(R(\theta)-R_{E,n}(\theta)\right)
>
b(n,\M_k)+\eta_k(n,y)
\right)
\\
&\leqslant
\pp\left(
\sup_{\theta\in\M_k}
\left(R(\theta)-R_{E,n}(\theta)\right)
>
b(n,\M_k)
\right.
\\
&\left.
\qquad+
\sqrt{
\frac{
2\left(
\sigma^2(\Theta)
+
2(L+R^*) b(n,\M_k)
\right)y
}{n}
}
+
\frac{(L+R^*)y}{3n}
\right)
\\
&\leqslant e^{-y}.
\endaligned
\nneq
Here, if $\M_k=\emptyset$, the corresponding probability is understood to be zero.

Taking $k=0$ and $y=x$ in \eqref{layer-bound}, we get
\bbeq\label{I0-bound}
\aligned
(\Pi_0)
&=
\pp\left(
\sup_{\theta\in\M_0}
\left(R(\theta)-R_{E,n}(\theta)\right)>\vep
\right)
\\
&=
\pp\left(
\sup_{\theta\in\M_0}
\left(R(\theta)-R_{E,n}(\theta)\right)
>
b(n,\M_0)+\eta_0(n,x)
\right)
\\
&\leqslant e^{-x}.
\endaligned
\nneq

For $1\leqslant k\leqslant M$ with $\M_k\neq\emptyset$, taking $y=kx$ in \eqref{layer-bound} yields
\bbeq\label{Ik-prebound}
\aligned
&\pp\left(
\sup_{\theta\in\M_k}
\left(R(\theta)-R_{E,n}(\theta)\right)
>
b(n,\M_k)+\eta_k(n,k x)
\right)
\leqslant e^{-kx}.
\endaligned
\nneq
By the assumption \eqref{thmM26bb} and the definition
$$
\vep=b(n,\M_0)+\eta_0(n,x),
$$
we have, for every $1\leqslant k\leqslant M$ with $\M_k\neq\emptyset$,
$$
\sqrt{k}\delta+\vep
=
\sqrt{k}\delta+b(n,\M_0)+\eta_0(n,x)
\geqslant
b(n,\M_k)+\eta_k(n,kx).
$$
Therefore, by \eqref{Ik-prebound},
\bbeq\label{Ik-bound}
\aligned
(\Pi_k)
&=
\pp\left(
\sup_{\theta\in\M_k}
\left(R(\theta)-R_{E,n}(\theta)\right)
>
\sqrt{k}\delta+\vep
\right)
\\
&\leqslant
\pp\left(
\sup_{\theta\in\M_k}
\left(R(\theta)-R_{E,n}(\theta)\right)
>
b(n,\M_k)+\eta_k(n,kx)
\right)
\\
&\leqslant e^{-kx}.
\endaligned
\nneq

Combining \eqref{I0-bound} and \eqref{Ik-bound} with the decomposition
\eqref{thmM26b}, we obtain
$$
p_+(n,\vep)
\leqslant
e^{-x}+\sum_{k=1}^{M}e^{-kx}
\leqslant
e^{-x}+\frac{e^{-x}}{1-e^{-x}}.
$$
This proves \eqref{p+estimateL}.
\nprf

\subsection{Proof of Theorem \ref{thmM39}}

\bprf[Proof of Theorem \ref{thmM39}]
For $L>0$, set
$$
\aligned
&Q_\theta^L:=(-L)\vee Q_\theta,
\qquad
&R^L(\theta):=\mu(Q_\theta^L),\\
&R_{E,n}^L(\theta):=L_n(Q_\theta^L),
\qquad
&R_*^L:=\inf_{\theta\in\Theta}R^L(\theta).
\endaligned
$$
Recall that
$$
\chi_L:=Q^*\mathbf 1_{\{Q^*>L\}}.
$$
By the definition of the $\psi_1$-norm, for every $t>0$, 
$$ 
\pp(Q^*>t) \leqslant 2\exp\left( -\frac{t}{\|Q^*\|_{\psi_1}} \right). 
$$ 
Therefore, 
\bbeq\label{tail-mean-bound} 
\aligned \mu(\chi_L) &= L\pp(Q^*>L) + \int_L^{+\infty}\pp(Q^*>t)\,dt \\
 &\leqslant 2\left( L+\|Q^*\|_{\psi_1} \right) \exp\left( -\frac{L}{\|Q^*\|_{\psi_1}} \right) \\
  &\leqslant \frac{\delta}{4}, 
  \endaligned 
  \nneq 
  where the last inequality follows from the condition \eqref{thmM39a}.

\smallskip
\noindent
{\bf Step 1. Truncation estimates and decomposition.}
By {\bf (H3)}, for every $\theta\in\Theta$,
\bbeq\label{trunc-diff}
0
\leqslant
Q_\theta^L-Q_\theta
\leqslant
\chi_L.
\nneq
Consequently,
\bbeq\label{trunc-risk-diff}
0
\leqslant
R^L(\theta)-R(\theta)
\leqslant
\mu(\chi_L),
\qquad
0
\leqslant
R_{E,n}^L(\theta)-R_{E,n}(\theta)
\leqslant
L_n(\chi_L).
\nneq
In particular,
\bbeq\label{eq520}
R_*
\leqslant
R_*^L
\leqslant
R_*+\mu(\chi_L),
\nneq
and
\bbeq\label{inf-trunc-emp}
0
\leqslant
\inf_{\theta\in\Theta}R_{E,n}^L(\theta)
-
\inf_{\theta\in\Theta}R_{E,n}(\theta)
\leqslant
L_n(\chi_L).
\nneq

By \eqref{eq520} and \eqref{inf-trunc-emp}, if
$$
R_*^L
\leqslant
\inf_{\theta\in\Theta}R_{E,n}^L(\theta)
+\frac{\vep}{2}
$$
and
$$
L_n(\chi_L)
\leqslant
\frac{\vep}{2},
$$
then
$$
\aligned
R_*
\leqslant
R_*^L
&\leqslant
\inf_{\theta\in\Theta}R_{E,n}^L(\theta)
+\frac{\vep}{2}
\\
&\leqslant
\inf_{\theta\in\Theta}R_{E,n}(\theta)
+
L_n(\chi_L)
+\frac{\vep}{2}
\\
&\leqslant
\inf_{\theta\in\Theta}R_{E,n}(\theta)
+\vep.
\endaligned
$$
Consequently,
$$
\aligned
&
\left\{
R_*>
\inf_{\theta\in\Theta}R_{E,n}(\theta)+\vep
\right\}
\\
&\subset
\left\{
R_*^L>
\inf_{\theta\in\Theta}R_{E,n}^L(\theta)
+\frac{\vep}{2}
\right\}
\cup
\left\{
L_n(\chi_L)>\frac{\vep}{2}
\right\}.
\endaligned
$$
Therefore, 
\bbeq\label{trunc-decomp}
\aligned
p_+(n,\vep)
&\leqslant
\pp\left(
R_*^L>
\inf_{\theta\in\Theta}R_{E,n}^L(\theta)
+\frac{\vep}{2}
\right)+ 
\pp\left(
L_n(\chi_L)>\frac{\vep}{2}
\right).
\endaligned
\nneq

It remains to estimate the truncated minimal-risk term and the empirical tail term.

\smallskip
\noindent
{\bf Step 2. The truncated minimal-risk term.}
We repeat the peeling argument in the proof of Theorem \ref{thmM26}
for the truncated class
$\{Q_\theta^L:\theta\in\Theta\}$
on the original risk-level sets
$\M_k$, $0\leqslant k\leqslant M$.

Since the truncation map
$u\mapsto(-L)\vee u$
is $1$-Lipschitz, for every
$\theta,\theta'\in\Theta$,
$$
\|Q_\theta^L-Q_{\theta'}^L\|_{\psi_1}
\leqslant
\|Q_\theta-Q_{\theta'}\|_{\psi_1}.
$$
Moreover, using the independent-copy representation of the variance, we have
$$
{\rm Var}_\mu(Q_\theta^L)
\leqslant
{\rm Var}_\mu(Q_\theta),
\qquad
\theta\in\Theta.
$$
Thus, on every non-empty set $\M_k$, the entropy bias is bounded by
$b(n,\M_k)$; namely,
\bbeq\label{trunc-bias-layer}
\ee\sup_{\theta\in\M_k}
\left(
R^L(\theta)-R_{E,n}^L(\theta)
\right)
\leqslant
b(n,\M_k).
\nneq

By \eqref{trunc-risk-diff} and \eqref{tail-mean-bound},
$$
R^L(\theta)
\leqslant
R(\theta)+\mu(\chi_L)
\leqslant
R^*+\frac{\delta}{4}.
$$
Since $Q_\theta^L\geqslant-L$, it follows that
\bbeq\label{trunc-upper-bound}
R^L(\theta)-Q_\theta^L
\leqslant
R^*+L+\frac{\delta}{4},
\qquad
\theta\in\Theta.
\nneq
Therefore, Bousquet's inequality gives, for every
$0\leqslant k\leqslant M$ such that $\M_k$ is non-empty and every $y>0$,
\bbeq\label{trunc-layer}
\pp\left(
\sup_{\theta\in\M_k}
\left(
R^L(\theta)-R_{E,n}^L(\theta)
\right)
>
b(n,\M_k)+\eta_{k,L}(n,y)
\right)
\leqslant
e^{-y}.
\nneq

For $k=0$, the definition of $\vep$ gives
$$
\frac{\vep}{2}
\geqslant
b(n,\M_0)+\eta_{0,L}(n,x).
$$

Since 
$$ 
R_*^L \leqslant \inf_{\theta\in\M_0}R^L(\theta), 
$$ 
we have 
$$ 
\aligned 
& \left\{ R_*^L> \inf_{\theta\in\M_0}R_{E,n}^L(\theta) +\frac{\vep}{2} \right\} \\ 
&\subset \left\{ \sup_{\theta\in\M_0} \left( R^L(\theta)-R_{E,n}^L(\theta) \right) > \frac{\vep}{2} \right\}. 
\endaligned 
$$ 
Hence, applying \eqref{trunc-layer} to $\M_0$ with $y=x$, we obtain 
\bbeq\label{trunc-I0} 
\pp\left( R_*^L> \inf_{\theta\in\M_0}R_{E,n}^L(\theta) +\frac{\vep}{2} \right) \leqslant e^{-x}. 
\nneq

For $1\leqslant k\leqslant M$ with $\M_k\neq\emptyset$,
\eqref{eq520} and \eqref{tail-mean-bound} imply
$$
\aligned
\inf_{\theta\in\M_k}R^L(\theta)-R_*^L
&\geqslant
\inf_{\theta\in\M_k}R(\theta)
-
R_*
-
\mu(\chi_L)
\\
&\geqslant
\sqrt{k}\delta-\frac{\delta}{4}.
\endaligned
$$
Hence
\bbeq\label{eq527}
\aligned
&\pp\left(
R_*^L>
\inf_{\theta\in\M_k}R_{E,n}^L(\theta)
+\frac{\vep}{2}
\right)
\\
&\leqslant
\pp\left(
\sup_{\theta\in\M_k}
\left(
R^L(\theta)-R_{E,n}^L(\theta)
\right)
>
\sqrt{k}\delta-\frac{\delta}{4}
+\frac{\vep}{2}
\right).
\endaligned
\nneq
By \eqref{thmM39-layer} and the definition of $\vep$,
$$
\aligned
\sqrt{k}\delta-\frac{\delta}{4}
+\frac{\vep}{2}
&\geqslant
b(n,\M_k)-b(n,\M_0)
+
\eta_{k,L}(n,kx)-\eta_{0,L}(n,x)
\\
&\quad+
b(n,\M_0)+\eta_{0,L}(n,x)
\\
&=
b(n,\M_k)+\eta_{k,L}(n,kx).
\endaligned
$$
Therefore, applying \eqref{trunc-layer} with $y=kx$, we obtain
\bbeq\label{trunc-Ik}
\pp\left(
R_*^L>
\inf_{\theta\in\M_k}R_{E,n}^L(\theta)
+\frac{\vep}{2}
\right)
\leqslant
e^{-kx}.
\nneq

Since $\Theta=\bigcup_{k=0}^{M}\M_k$, with empty sets omitted,
$$
\aligned
&
\left\{
R_*^L>
\inf_{\theta\in\Theta}R_{E,n}^L(\theta)
+\frac{\vep}{2}
\right\}
\\
&\subset
\bigcup_{k=0}^{M}
\left\{
R_*^L>
\inf_{\theta\in\M_k}R_{E,n}^L(\theta)
+\frac{\vep}{2}
\right\}.
\endaligned
$$
Combining the estimates over the non-empty sets
$\M_k$, $0\leqslant k\leqslant M$, gives
\bbeq\label{trunc-pplus-eps}
\aligned
&\pp\left(
R_*^L>
\inf_{\theta\in\Theta}R_{E,n}^L(\theta)
+\frac{\vep}{2}
\right)
\\
&\leqslant
e^{-x}
+
\sum_{k=1}^{M}e^{-kx}
\\
&\leqslant
e^{-x}
+
\frac{e^{-x}}{1-e^{-x}}.
\endaligned
\nneq

\smallskip
\noindent
{\bf Step 3. The empirical tail term.}
Since
$\chi_L\leqslant Q^*$
and
$\|Q^*\|_{\psi_1}<+\infty$,
we have
$\|\chi_L\|_{\psi_1}<+\infty$.
Applying Theorem \ref{thm_BV} together with
\eqref{Bern-c} to the single observable $\chi_L$, we obtain
$$
\pp\left(
L_n(\chi_L)-\mu(\chi_L)
>
2\|\chi_L\|_{\psi_1}\sqrt{\frac{x}{n}}
+
\|\chi_L\|_{\psi_1}\frac{x}{n}
\right)
\leqslant
e^{-x}.
$$
Equivalently,
\bbeq\label{tail-bern}
\pp\left(
L_n(\chi_L)>
\tau_n(x,L)
\right)
\leqslant
e^{-x}.
\nneq
By \eqref{vep12},
\bbeq\label{thmM39e}
\tau_n(x,L)
\leqslant
\frac{\vep}{2}.
\nneq
Consequently,
\bbeq\label{tail-small}
\pp\left(
L_n(\chi_L)>
\frac{\vep}{2}
\right)
\leqslant
e^{-x}.
\nneq

\smallskip
\noindent
{\bf Step 4. Conclusion.}
Combining
\eqref{trunc-decomp},
\eqref{trunc-pplus-eps}
and
\eqref{tail-small},
we obtain
$$
p_+(n,\vep)
\leqslant
2e^{-x}
+
\frac{e^{-x}}{1-e^{-x}}.
$$
This proves \eqref{vep11}.
\nprf

\section{Appendix}

In this section, we will specify the absolute constants $K_1, K_2$ mentioned in Theorem \ref{lem41}. The proof is based on van der Vaart and Wellner's methods with a new technique. First, let us recall a lemma which is needed in the proof.

\blem[Lemma 2.2.1 \cite{vdVW96}]\label{lem221}
Let $X$ be a random variable with $\mathbb{P}(|X| >x)\leqslant Be^{-Cx^p}$ for every $x$, for constants $B$ and $C$, and for $p \geq 1$. Then its Orlicz norm satisfies $\| X\|_{\psi_p}\leqslant ((1+B)/C)^{\frac{1}{p}}$.
\nlem

\blem\label{ipconst}
  If $\max_{1 \leqslant i \leqslant m}\|  X_i\|_{\psi_p} \leqslant A$, where $A$ is a positive constant, then
  \begin{equation}\label{improvedconst}
  \| \max_{1 \leqslant i \leqslant m} |X_i| \|_{\psi_p} \leqslant A \left(\frac{\log(1+m)}{\log 2}\right)^{\frac{1}{p}}.
  \end{equation}
\nlem

\begin{proof}[Proof of Lemma \ref{ipconst}] By the definition of Orlicz norm, $\| X_i \|_{\psi_p} \leqslant A$ implies
$$
\mathbb{E} [\exp((|X_i|/A)^p)-1]\leqslant 1.
$$

Let $Y:= \exp((\max_{i}|  X_i|/A)^p)-1$. Then, $$Y=\max_i(\exp((|X_i|/A)^p)-1) \leqslant \sum_{i=1}^{m}[\exp((|X_i|/A)^p)-1], $$
which leads to

  $$
  \mathbb{E}\exp((\max_{i}|  X_i|/A)^p)=\mathbb{E}Y+1 \leqslant 1+m.
  $$

  For $\alpha \in (0, 1]$, by Jensen's inequality,
  $$
  \mathbb{E}\exp(\alpha(\max_{i}|  X_i|/A)^p) \leqslant [\mathbb{E}\exp((\max_{i}|  X_i|/A)^p)]^{\alpha} \leqslant (1+m)^{\alpha}.
  $$

  Set $\alpha =\frac{\log 2}{\log (1+m)}$ which belongs to $(0, 1]$,
  $$
  \mathbb{E}[\exp(\alpha(\max_{i}|  X_i|/A)^p)-1] \leqslant (1+m)^{\alpha}-1=1.
  $$

  Therefore,
  $$
  \| \max_{i} |X_i| \|_{\psi_p} \leqslant A\alpha ^{-\frac{1}{p}} =A\left(\frac{\log (1+m)}{\log 2}\right)^{\frac{1}{p}}.
  $$

\end{proof}

Incorporating lemma \ref{ipconst} into van der Vaart and Wellner's results \cite[Proof of Lemma 2.2.10]{vdVW96} yields a new proof of Theorem \ref{lem41}.

\begin{proof}[Proof of Theorem \ref{lem41}]

The proof is divided into two steps.

{\bf Step 1.} The condition implies the upper bound $2\exp(-x^2/(4c_B))$ on $\mathbb{P}(|X_i| > x)$ for $x \leqslant \frac{c_B}{M}$, and the upper bound $2\exp(-x/(4M))$ for other positive $x$. Consequently, the same upper bounds hold for all $x >0$ for the probabilities $\mathbb{P}(|X_i|\mathbf 1_{\{|X_i|\leqslant c_B/M\}}>x)$ and $\mathbb{P}(|X_i|\mathbf 1_{\{|X_i|> c_B/M\}}>x)$, respectively.

By Lemma \ref{lem221}, this implies that
$$
\| |X_i|\mathbf 1_{\{|X_i|\leqslant c_B/M\}}\|_{\psi_2}\leqslant ((1+2)4c_B)^{\frac{1}{2}}=\sqrt{12c_B}
$$
and
$$
\| |X_i|\mathbf 1_{\{|X_i|> c_B/M\}}\|_{\psi_1}\leqslant ((1+2)4M)=12M.
$$

Also, from Jensen's inequality we have for $1 \leqslant p \leqslant q$ and for
any r.v. $X$ (see \cite[p. 95 and Problem 2.2.5]{vdVW96} for a similar result),
\bbeq\label{prob5}
\| X\|_{\psi_p} \leqslant \| X\|_{\psi_q}(\log 2)^{\frac{1}{q}-\frac{1}{p}}.
\nneq
In fact, let the concave function $\phi$ satisfy that $\phi(\psi_q((\log 2)^{1/q}x))=\psi_p(x(\log
 2)^{1/p})$ and $\phi(1)=1$ for $q \geqslant p
 \geqslant 1$. Hence, $\phi$ is nondecreasing and
 $$
 \aligned
 \phi\left( \ee \psi_q\left(\frac{|X|(\log 2)^{1/q}}{\|X \|_{\psi_p}(\log 2)^{1/p}}\right)\right)
  &\geqslant \ee\phi\left(  \psi_q\left(\frac{|X|(\log 2)^{1/q}}{\|X \|_{\psi_p}(\log 2)^{1/p}}\right)\right)\\
  &= \ee \psi_p\left(\frac{|X|(\log 2)^{1/p}}{\|X \|_{\psi_p}(\log 2)^{1/p}}\right)\\
  &=1,
 \endaligned
 $$
which implies \eqref{prob5}.

Taking $p=1$ and $q=2$, we get $\| X\|_{\psi_1} \leqslant \| X\|_{\psi_2}(\log 2)^{-\frac{1}{2}}$.

{\bf Step 2.} By Lemma \ref{ipconst}, we get
$$
\|\max_i |X_i|\|_{\psi_2} \leqslant \left(\frac{\log(1+m)}{\log2}\right)^{\frac{1}{2}}\max_i\|X_i\|_{\psi_2},
\\
\|\max_i |X_i|\|_{\psi_1} \leqslant \frac{\log(1+m)}{\log2}\max_i\|X_i\|_{\psi_1}.
$$

Consequently,
\begin{align*}
\| \max_i |X_i| \|_{\psi_1}
&\leqslant \| \max_i |X_i|\mathbf 1_{\{|X_i|\leqslant c_B/M\}}\|_{\psi_1}+\| \max_i |X_i|\mathbf 1_{\{|X_i|> c_B/M\}}\|_{\psi_1}\\
&\leqslant \frac{1}{\sqrt{\log 2}}\| \max_i |X_i|\mathbf 1_{\{|X_i|\leqslant c_B/M\}}\|_{\psi_2}+\| \max_i |X_i|\mathbf 1_{\{|X_i|> c_B/M\}}\|_{\psi_1}\\
& \leqslant \frac{1}{\sqrt{\log 2}}\left(\frac{\log(1+m)}{\log2}\right)^{\frac{1}{2}}\max_i\||X_i| \mathbf 1_{\{|X_i|\leqslant c_B/M\}}\|_{\psi_2}\\
&\quad+\frac{\log(1+m)}{\log2}\max_i\||X_i| \mathbf 1_{\{|X_i|> c_B/M\}}\|_{\psi_1}\\
&\leqslant \frac{2\sqrt{3}}{\log 2}\sqrt{c_B\log(1+m)}+\frac{12}{\log2} M\log(1+m),
\end{align*}
which implies that $K_1=\frac{2\sqrt{3}}{\log 2}, K_2=\frac{12}{\log2}.$
\end{proof}

\bigskip

{\bf Acknowledgements.} We are grateful to Arnaud Guillin for the communication of the recent reference \cite{Esc23}. 
Part of this joint work was carried out while the second-named author was a visiting Ph.D. student at LMBP, Universit\'e Clermont-Auvergne,
from July 2024 to July 2025. He expresses his gratitude to LMBP and to Prof.
H. Djellout for the warm hospitality.

\bigskip
\bibliographystyle{plain}

\end{document}